\documentclass[12pt]{article}
\usepackage{blindtext}
\usepackage[dvips]{graphicx}
\usepackage{amsfonts,epsfig,cite,array,multirow,graphicx,amsmath,ltablex,tabularx,setspace,arydshln,amssymb,multirow}
\usepackage{subfigure,calc,amsmath,amstext,amsthm,pslatex,multicol}
\usepackage{colortbl,graphics,url}
\usepackage{textcomp}
\usepackage{algorithm}
\usepackage{algpseudocode}
\usepackage{enumitem}
\usepackage{booktabs}
\usepackage{bigstrut}
\usepackage{mathrsfs}

\makeatletter
\newcommand{\removelatexerror}{\let\@latex@error\@gobble}
\makeatother

\providecommand{\keywords}[1]{\textbf{\textit{Index terms---}} #1}

\begin{document}

\title{An Adaptive Training-less System for Anomaly Detection in Crowd Scenes}

\author{Arindam~Sikdar~and~Ananda~S~Chowdhury
\thanks{The authors are with the Department
of Electronics and Telecommunication Engineering, Jadavpur University, Kolkata - 700032, India.}
\thanks{E-mails: arindamsikdarju@gmail.com, as.chowdhury@jadavpuruniversity.in}}


\maketitle

\begin{abstract}
Anomaly detection in crowd videos has become a popular area of research for the computer vision community. Several existing methods generally perform a prior training about the scene with or without the use of labeled data. However, it is difficult to always guarantee the availability of prior data, especially, for scenarios like remote area surveillance. To address such challenge, we propose an adaptive training-less system capable of detecting anomaly on-the-fly while dynamically estimating and adjusting response based on certain parameters. This makes our system both training-less and adaptive in nature. Our pipeline consists of three main components, namely, adaptive 3D-DCT model for multi-object detection-based association, local motion structure description through saliency modulated optic flow, and anomaly detection based on earth movers distance (EMD). The proposed model, despite being training-free, is found to achieve comparable performance with several state-of-the-art methods on the publicly available UCSD, UMN, CHUK-Avenue and ShanghaiTech datasets.\\
\end{abstract}


%
\section{Introduction}
\label{sec_Intro} 
Anomaly detection in crowd videos has evolved as an important research problem for the computer vision community. Unavailability of surveillance data in low resolution form, paucity of sufficient training examples, especially for rare, sparse and anomalous events in a video, and computational burden for training have made this problem immensely challenging. One common approach to solve this problem is to develop a structure-aware crowd model~\cite{Mehran_2009,Yuan_Yuan_2015}, which efficiently discriminates between the normal and abnormal patterns in crowd scenes. However, most of the approaches assume an underlying (particle) interaction model in addition to being dependent on labeled training samples \cite{Mehran_2009,Yuan_Yuan_2015,Wu_2010}. It is often not wiser to rely on such assumption as the mechanics of human crowd possesses both dynamic and psychological characteristics~\cite{Ali:2008}, resulting in a particular set of motion pattern~\cite{Bera_2016_CVPR_Workshops,Li_2015} which is quite different from particle interactions in constrained space. More importantly, availability of training data cannot be guaranteed in scenarios like remote area surveillance. Therefore, a model is necessary, which can invariably handle diverse scenes without training and still able to capture the crowd structure patterns without relying on any particle interaction assumption. We term such models as \textquotedblleft \textbf{training-less}\textquotedblright{} models. Although there are some works~\cite{Del_Giorno_2016} in the literature that have accurately handled anomaly detection without using labeled data, they still trained classifier based on on distributions as prior to determine anomaly.\\
\begin{figure}[t!]
\centering
\includegraphics[width=.95\textwidth]{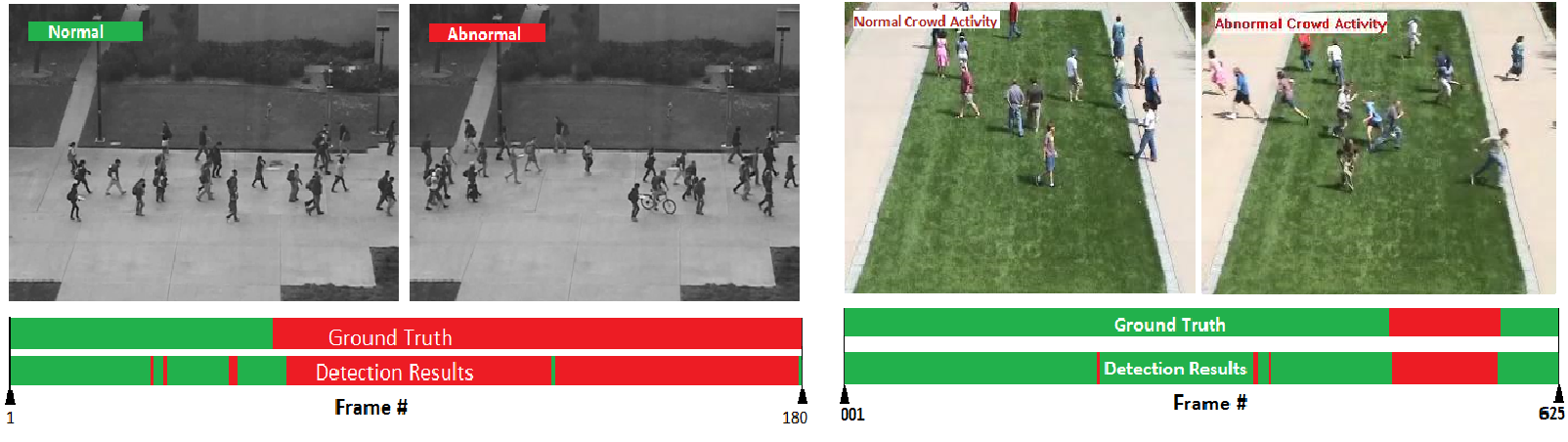}
\caption{Anomaly detection in UCSD and UMN datasets using the proposed method. The length of the color bar indicates the duration of the videos where green and red region indicates non-anomalous region and anomalous regions respectively.}
\label{fig:figure1}
\end{figure}
\indent In sharp contrast, the proposed adaptive training-less system can run on-the-fly for anomaly detection. In figure~\ref{fig:figure1}, we show one scene from the UCSD and three scenes from the UMN datasets which are marked as normal or abnormal using our solution. The basic strategy of detecting anomaly is to first break the complex motion dynamics into simpler ones as local motion structures/descriptors and then analyze the cumulative effect of such structures indicating the occurrence of an anomaly. Since our model is training-free one obvious question is that how it can handle non-stationary nature of abnormality? The answer lies in the fact that our model dynamically determines the possible target regions in form of proposals and construct descriptors only on those regions to determine abnormality. Later in the paper, in figure \ref{fig:figure18}, we explicitly demonstrate how our method can successfully handle non-stationary abnormality. Owing to the stochastic nature of the anomaly it is in fact quite difficult to train a specific model with a specific mode of anomaly. A self-adjustable training-less system is more preferable instead.
\begin{figure*}[t]
\centering
\includegraphics[width=0.9\textwidth]{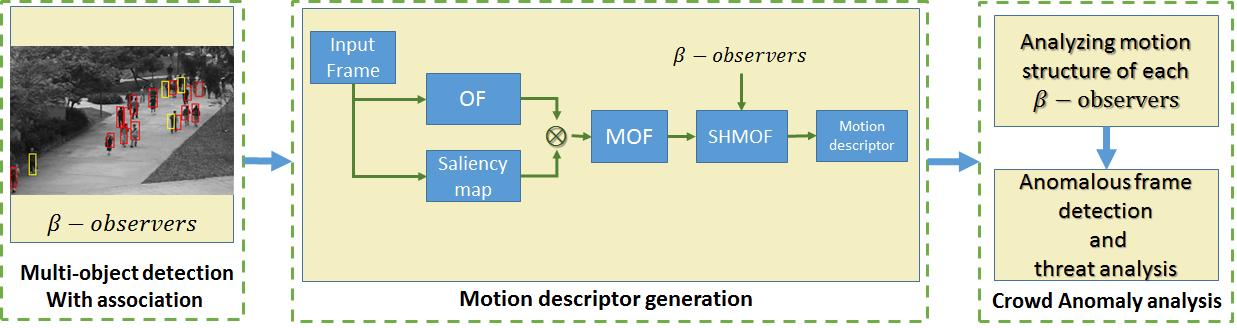}
\caption{A schematic of our proposed framework composed of 3 major components.}
\label{fig:figure2}
\end{figure*}
\section{Related Work}
\label{sec_priorWorks}
Several works were reported on crowd anomaly detection ~\cite{Sodemann_2012,Xuan_Mo_2014}. The methods developed so far can be broadly divided into two major groups, namely, \textit{trajectory based approaches} and \textit{object representation based approaches}. \\ \indent The first category of methods are based on intuitive segmentation and exhaustive tracking of objects/individuals in a scene based on trajectory based models~\cite{Bera_2016_CVPR_Workshops,Wu_2010,Zhang_2013,Stauffer_2000,Cui_2011,Piciarelli_2008}. 
In this category, a model is trained with normal set of trajectories and any abrupt deviation of those trajectories classifies an anomaly. 
Key point based trajectory analysis were adopted by Wu~\textit{et al.}~\cite{Wu_2010} where each representative point was treated as a chaotic Lagrangian particle and was tracked by utilizing invariant feature descriptors. Cui~\textit{et al.}~\cite{Cui_2011} explicitly exploited the relationships among a group of people by extracting normal/abnormal patterns using interaction energy potentials (IEP) and SVM classifier. A recent work was reported on online crowd behavior learning by Bera \textit{et al.}~\cite{Bera_2016_CVPR_Workshops}. They analyzed pedestrian behavior by combining nonlinear pedestrian models with Bayesian learning at trajectory-level. Though they categorized anomalies into low, medium and high threat levels, no detailed analysis was performed. Zhang \textit{et al.}~\cite{Zhang_2013} modeled crowd event using Bag of Trajectory Graphs (BoTG) and classified crowd anomaly using mean shift~\cite{Comaniciu_2002} trajectory clustering approach. In-spite of having explicitly high-semantics on interpreting abnormality, these methods are often infeasible in terms of exhaustive tracking of each individual in crowded scenes. These methods are also computationally quite expensive for keeping traces of long-duration trajectories. 
\\ \indent The second group of methods \cite{Adam_2008,Ali_2007,Mehran_2009,Thida_2013}  do not involve tracking. These methods mostly include motion and appearance modeling using pixel/blob changes \cite{Benezeth_2009,Saligrama_2012,Kratz_2009,Wu_2014} or optical flow variations~\cite{Ali_2007,Adam_2008}. Ali and Shah~\cite{Ali_2007} succeeded in motion pattern segmentation in changing crowd scenes using finite time Lyapunov exponent (FTLE) field and extended their framework to anomaly detection. Kim and Grauman~\cite{Kim_Grauman_2009} employed probabilistic PCA to model local optical flow patterns and enforced global consistency using Markov Random Field (MRF). Cong \textit{et al.}~\cite{Cong_2013} represented motion pattern in image sequence using multiscale histogram of optical flow. For this method unusual events are identified with high reconstruction cost/error of a pre-trained sparse model. Mehran~\textit{et al.}~\cite{Mehran_2010} found an analogy between dynamics of crowd flow and fluid flow and employed it to represent abnormal behavior and traffic dynamics. Some models have drawn inspiration from classical studies and portrayed flow with interactive features such as social force models (SFM)~\cite{Ali_2007,Mehran_2009}. In~\cite{Mehran_2009}, Latent Dirichlet allocation (LDA) have been employed to explicitly distinguish normal/abnormal motion patterns. Su~\textit{et al.}~\cite{Su_2013} also employed LDA to recognize crowd behaviors using a codebook built from spatio-temporal features. Motion and spatial information have also been modeled either independently or jointly in many frameworks such as~\cite{Benezeth_2009,Kim_Grauman_2009,Kratz_2009,Zhang_2005}. In some works~\cite{Kratz_2009,Zhang_2005}, authors have shown temporal consistency of the normal temporal sequences using hidden Markov models (HMMs) while~\cite{Benezeth_2009} and~\cite{Kim_Grauman_2009} relied on Markov Random Field (MRF) to enforce global spatial consistency. Spatio-temporal variations were learned for detection of abnormal regions in crowd video using Laplacian eigenmaps by Thida~\textit{et al.}~\cite{Thida_2013}. In~\cite{Mahadevan_2010}, the authors modeled crowd behavior using dynamic textures (DTs). Li \textit{et al.} \cite{Weixin_Li_2014} used dynamic texture and center-surround saliency detector for anomaly detection. In \cite{Wu_2014}, the authors employed Bayesian models for crowd anomaly detection. 

\indent For both the categories of methods, one common limitation is that their success depends on the availability of labeled dataset (normal or abnormal patterns) for training. Another shortcoming of these methods is that no additional motion information is used for improving flow based descriptors. A close comparison of our training-less system would be with the work of Del~\textit{et al.}~\cite{Del_Giorno_2016} where they have detected crowd anomaly using unlabeled data and measured abrupt changes in a video. Although, they have not used any labeled training data, but they still generated labels based on change in distribution to explicitly train a logistic regression model. In our proposed training-less system, parameter learning/adjustment occurs on-the -fly with no explicit model training. We now state our contributions:
\begin{enumerate}
\item The major contribution lies in developing a training-less system of crowd anomaly analysis. Even without a training phase, we have achieved comparable performance with several state-of-the-art methods.  
\item On the theoretical side, we have modulated the optical flow (OF) map by a temporal saliency map through a reward-penalty function before generating local descriptors. Saliency by itself is difficult to be associated with anomaly. However, it can act as a cue to OF map to significantly improve the model performance by yielding contrastive motion descriptor from which anomaly is eventually determined.
\item Finally, we have developed an adaptive 3D-DCT model that works on a self-adjustable scaling parameter for pedestrian detection. 
\end{enumerate}
\begin{figure*}[t!]
\centering
\includegraphics[width=\textwidth]{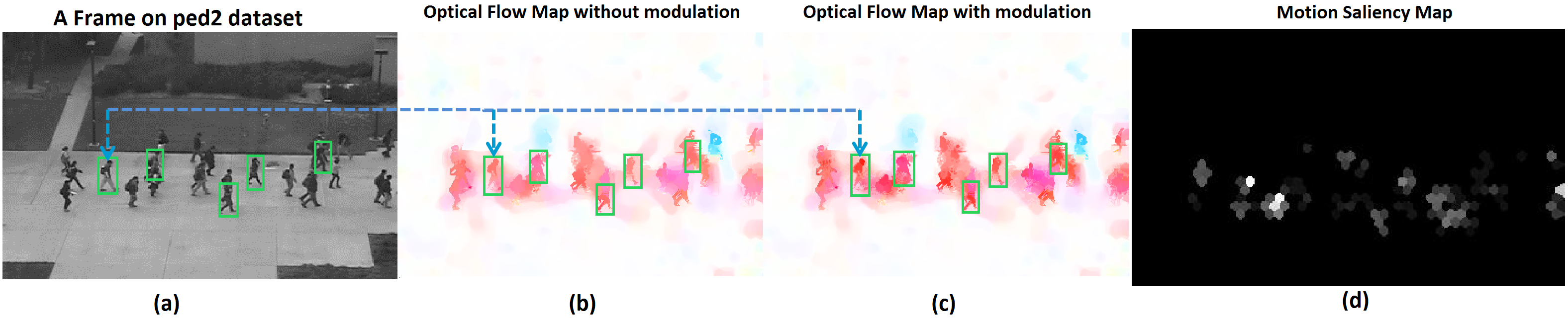}
\caption{An example of modulation in OF map represented in Middlebury color coding with (a) an frame/instant on ped2 dataset, (b) OF without modulation, (c) OF with modulation, (d) temporal saliency map. The figure illustrates the region belonging to pedestrian within bounding boxes (green boxes) that have undergone enhancement over suppression based on the motion saliency.}
\label{fig:figure5}
\end{figure*}
\section{Proposed Framework}
\label{sec_framework}
In this section, we provide a detailed description of our proposed method. The three major components of our solution are: A) Multi-object association-based detection, B) Local structural descriptor generation, and C) Crowd anomaly analysis. A schematic diagram of the overall pipeline is shown in figure~\ref{fig:figure2}.
\subsection{Multi-object association-based detection}
\label{sec:secA} Multi-object detection-based association is composed of three major steps. These steps are discussed below in details. a) Generation of proposal bounding boxes, b) Adaptive 3D-DCT based association, and c) Object preservation  via Template pool building.  
\subsubsection{Generation of proposal bounding boxes}
\label{lab:bbox_generation}
In our previously published work~\cite{Sikdar_2017}, we have shown pedestrian detection is achievable in a training-less manner with certain restrictions. Each such pedestrian is an object in a frame which is finally localized using a bounding box. Before assigning a bounding box to an object we first separate the overall foreground region (region of motion) from the background using Gaussian mixture model (GMM). A foreground mask (region) is generated in colase for more than two objects if objects are spatially close or overlapped. To correctly localize such objects using separate bounding boxes, the mask/region first needs to be segmented. We have used edge-fused segmentation for that purpose. Accurate region segmentation for an object is hard to ensure as we have no prior information of object boundaries, moreover, there is also a chance of over-segmentation as edges occurs within an object. Thus we generate several proposal bounding boxes of varying size over centroid of these segments. Out of multiple proposals the bounding box that best localize an entire object is eventually preserved by 3D-DCT based association. The generation of a proposal bounding box BB~\cite{Sikdar_2017} is based on two factors, namely, area ($\phi^{BB}$) and entropy ($\psi^{BB}$). For a proposal bounding box $BB$, if its $\phi^{BB}$ and $\psi^{BB}$ are greater than thresholds, \textit{i.e.}, $\phi_{thres}$ and $\psi_{thres}$ respectively is considered as a valid proposal.

\begin{algorithm}[ht!]
\caption{Multi-Object detection with association}
\begin{algorithmic}[1]
\Require Video Sequence
\Ensure  $\beta$
\Statex \textit{Initilization:} $\phi_{thres}\leftarrow 0.5,~\psi_{thres}\leftarrow 0.7$
\For {t = 1 \textbf{to} T} /* $T$ is the total no. of frames */
\State Obtain foreground region using GMM
\State Perform edge-fused region segmentation
\State Generate proposal bounding boxes satisfying $\phi^{BB} > \phi_{thres}$ and $\psi^{BB}> \psi_{thres}$   
\If {$t==1$}
\State Apply NMS over $N$ templates
\State Initiate template pools based on equation~\ref{equ:equ_16}
\Else
\State Compute $Q^{BB}$ score based on equation~\ref{equ:equ_13}
\State Apply NMS over $N$ templates
\State Update template pools based on equation~\ref{equ:equ_15} and \ref{equ:equ_16}    
\EndIf
\State Obtain $\beta$ observers
\EndFor
\end{algorithmic}
\label{algo:algo1}
\end{algorithm}
\subsubsection{Adaptive 3D-DCT based association}
\label{sec:3D-DCT filtering}
We design an alternative version of 3D-DCT tracker~\cite{Li_2013} for associating multiple objects in an unsupervised manner. The model preserves a 3D-volume (template pool) of object bounding boxes (templates) for every object in an incremental and restricted manner. This set of image stacks are updated iteratively and utilized for associating relevant objects in every frame. The association is based on a likelihood value based quality score  scaled over an estimated parameter ($\xi$). This parameter is auto-adjusted at every frame which validates the adaptive nature of our model. Based on this scaling, a discriminative quality score is generated for each cross-association. Thus a template that best associates with a given template pool is considered its desirable associate. 
\begin{figure}[t!]
\begin{center}$
\begin{array}{cc}
\includegraphics[width=0.53\textwidth]{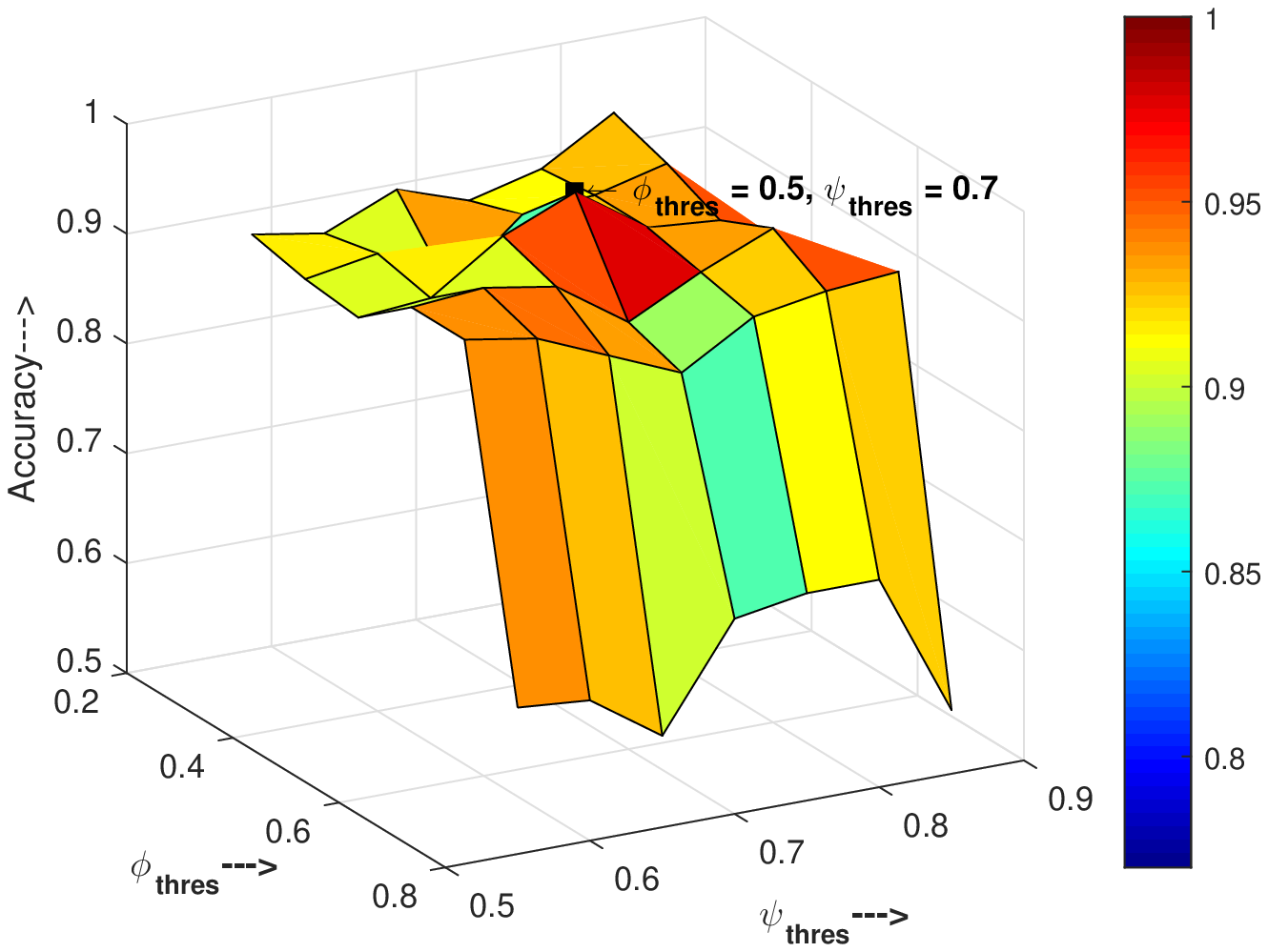}&
\includegraphics[width=0.47\textwidth]{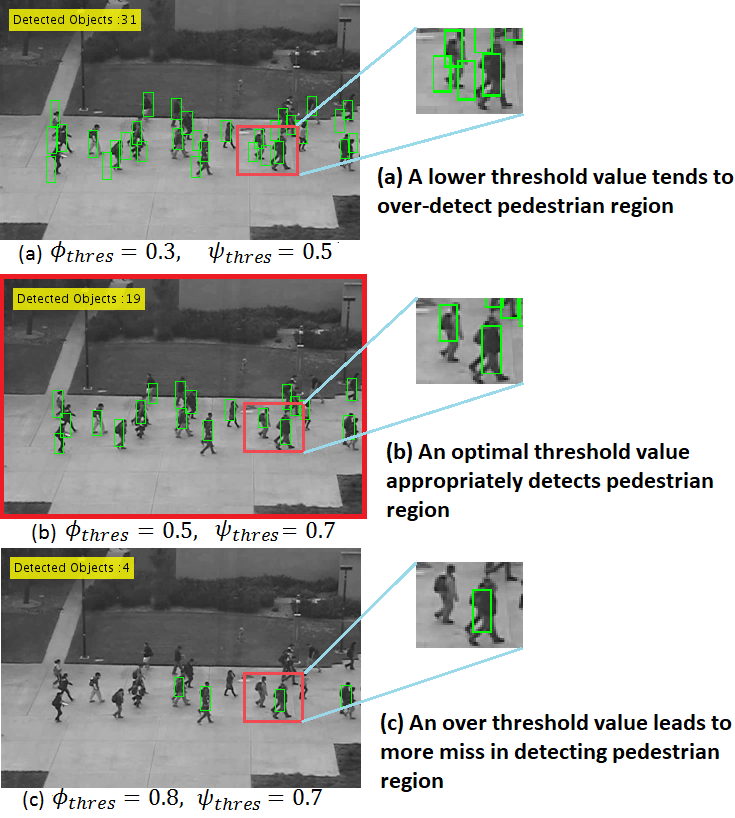}
\end{array}$
\end{center}
\centering
\caption{A typical example showing the effect multi-object detection performance over the threshold parameters $\phi_{thres}$ and $\psi_{thres}$ in terms of accuracy defined as (TP+TN)/(TP+TN+FP+FN) along with a frame showing the detection results over the optimal $\phi_{thres}$ and $\psi_{thres}$ selected.} 
\label{fig:figure9}
\end{figure}
Suppose there are $N$ proposal bounding boxes and $f$ objects over which a 3D-volume is required to build over a span of time $\Delta T$. For each object $\kappa \in f$, a 3D-volume is built from 2D-templates of an object also termed as template pool. Let each template pool built at time instant $t$ be denoted by $\{S_{\mathbf{III,\kappa}}(x,y,t)\} \in \mathbb{R}^{W \times H \times N_{temp}}$ where, $(x,y,t)$ are the coordinates of the 3D volume and $W, H, N_{temp}$ respectively denote the width, height and number of templates of an individual $\kappa$. The corresponding 3D-DCT coefficient volume of $S_{\mathbf{III,\kappa}}$ is denoted by $C_{\mathbf{III,\kappa}}$ and is computed as in~\cite{Li_2013}. We can thus recompute a low frequency approximation of $S_{\mathbf{III,\kappa}}$ as $S^*_{\mathbf{III,\kappa}}$ by first setting the coefficients of higher frequency component in $C_{\mathbf{III,\kappa}}$ to $0$ and then performing an inverse 3D-DCT transformation~\cite{Li_2013}. The coefficients corresponding to higher frequencies are usually sparse and mostly consist of texture clues while the low frequency coefficients are relatively dense and thus retained. Based on the loss of high frequency components, a reconstruction error is computed for newly arrived template $n_{\kappa}$ for object $\kappa$. The low frequency approximation of template $n_{\kappa}$ is the end 2D-template of updated $S_{\mathbf{III,\kappa}} \in \mathbb{R}^{W \times H \times (N_{temp} + 1)}$, \textit{i.e.}, $S^*_{\mathbf{III,\kappa}}(:,:,N_{temp}+1)$.  Likelihood measure of a template $n_{\kappa}$ is computed based on the reconstruction error $\parallel n_\kappa - S^*_{\textbf{III}}(:,:,N_{temp}+1) \parallel^2$ and is given by the following equation:  
\begin{equation}
\mathscr{L}(n_\kappa,\xi) = exp\Bigl(-\frac{1}{2\xi}\parallel n_\kappa - S^*_{\mathbf{III, \kappa}}(:,:,N_{temp}+1) \parallel^2 \Bigr)
\label{equ:equ_11}
\end{equation}
where, $\xi$ is the scaling factor. Out of many templates $N$, $n_{\kappa} \in N_{\kappa} \subset N$ such that set $N_{\kappa}$ has non-zero spatial overlap with the previous template of $S_{\mathbf{III,\kappa}}$, i.e., $S_{\mathbf{III,\kappa}}(:,:,N_{temp})$. Thus we can now define the set $N_\kappa = \{n_\kappa: area(n_\kappa \cap S_{\mathbf{III,\kappa}}(:,:,N_{temp})) \neq 0\}$, where, $area(\cdot)$ denotes the spatial overlap of $n_\kappa$ and $S_{\mathbf{III,\kappa}}(:,:,N_{temp})$ in between consecutive video frames. The likelihood function of equation~\ref{equ:equ_11} is normalized over the sigmoid function $\rho(x) = 1/(1+e^{-x})$ that lies within the range [0,1].
To properly select a template we make the normalized likelihood score for each $N_\kappa$ more discriminative based on the scaling parameter $\xi$. The scaling parameter is adaptively optimized over all such set of $N_{\kappa}$ for all objects $f$. An optimized $\xi$ will maximize the spread of likelihood values over sigmoid function $\rho(x)$. The optimal $\xi_{\kappa}$ is given by:
\begin{equation}
   \xi_{\kappa} = \underset{\xi}{\mathrm{argmax}}\{var(\rho(\mathscr{L}(n_\kappa,\xi)))\} 
    \label{equ:equ_12a}
\end{equation}
where, $var(\cdot)$ is the variance. Thus the parameter $\xi_{\kappa}$ becomes adaptive as its computation is based on the selected valid sets $N_{\kappa \in f}$. This is much preferred over a fixed $\xi$ as setting a arbitrary smaller value will unduly concentrate the normalized likelihood values towards 1, while an arbitrary high value of $\xi$ will concentrate all the values towards 0.5. The mean optimal $\hat{\xi}$ is computed from the set of optimal $\xi_\kappa$ of each $\kappa^{th}$ template pool in the following manner:
\begin{equation}
\hat{\xi} = \frac{1}{f}\sum^{f}_{\kappa=1}\xi_{\kappa}
\label{equ:equ_12b}
\end{equation}
To further enhance the multi-object association, we define the quality score of the $q^{th}$ template $n_q \in \underset{\kappa}{\bigcup}N_{\kappa}$, in the following manner: 
\begin{equation}
Q^{BB}_{q,\kappa} = \phi^{BB}_{q} * \underset{\kappa}{\mathrm{max}}\{ \rho(\mathscr{L}(n_q,\hat{\xi}))\}
\label{equ:equ_13}
\end{equation}
In the above equation, the factor $\phi^{BB}_{q}$ denotes the normalized entropy of the sample $n_q$ and the factor $\underset{\kappa}{\mathrm{argmax}}\{ \rho(\mathscr{L}(n_q,\hat{\xi}))\}$ denotes the maximum likelihood score (or least reconstruction error) when this template is compared with all the available template pools. The term $Q^{BB}_{q,\kappa}$ is a measure of the strength of association of the template $n_q$ with the $\kappa^{th}$ template pool. To keep the best matches between the competing overlapping templates, we apply non-maximal suppression (NMS) based on quality score $Q^{BB}_{q,\kappa}$. This quality score acts as confidence score in determining the best bounding box (template) for a particular template $\kappa$. 
\begin{figure}[t]
\centering
\includegraphics[width=0.55\textwidth]{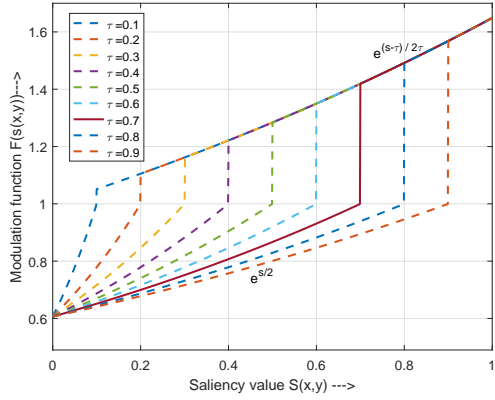}
\caption{A variation of modulating function $\mathcal{F}(s)$ over different values of $\tau$. Here, $\tau$ demarcates the range of saliency score for which the corresponding OF value at each pixel is enhanced or suppressed.}
\label{fig:figure6}
\end{figure}    
\subsubsection{Object preservation via Template pool building} 
\label{sec:template update}
Let NMS in the frame $t$ keeps $N'$ out of a total of $N$ templates. Out of these $N'$ templates, say, $\beta$ templates are associated with the previous template pools and the remaining $(N' - \beta)$ template pools are newly created. These $\beta$ template pools act as observers and the rest act as targets. From the total number of existing $f$ template pools, $(f - \beta)$ template pools are discarded if they are not associated for the past $\Delta T$ frames. Possible reasons behind no association include occlusion, exit, miss and false detection. Rejection of template pools reduces the possibility of over-fitting (false associations). It further restricts the number of template pools which would otherwise increase unnecessary computational overhead. Now, we discuss our strategy for template pool association.    

\indent  A template at time $t$, \textit{i.e.}, $n^t_{\kappa}$ updates a previously existing template pool $\mathcal{T}^{t-1}_\kappa$ after being associated with it. The updated template pool can be expressed as:
\begin{equation}
\begin{matrix}
\mathcal{T}^{t}_\kappa  = \{Concat(\mathcal{T}^{t-1}_\kappa,n^{t}_\kappa ): S_{\textbf{III}}(x,y,t-1)_{N_1 \times N_2 \times N_3} \longrightarrow \\ S_{\textbf{III}}(x,y,t)_{N_1 \times N_2 \times (N_3+1)}, \kappa \in \beta\}
\end{matrix}
\label{equ:equ_15}
\end{equation}
where, $Concat(\cdot)$ concatenates the newly arrived template. On the other hand, a new template pool created in the present frame can be expressed as:
\begin{equation}
\mathcal{T}^{t}_\kappa = \{\mathcal{T}^{t}_\kappa : S_{\textbf{III}}(x,y,t)_{N_1 \times N_2 \times 1}, \kappa \in (q'-\beta)\}
\label{equ:equ_16}
\end{equation}
So, the overall template pool $\mathcal{T}^{t}$ in the present frame $t$ can be represented as a union of two terms (which themselves are expressed as unions):\\ $\bigl( \underset{\kappa \in \beta}{\bigcup} \mathcal{T}^{t}_\kappa\bigr) \bigcup \bigl(\underset{\kappa \in (N' - \beta)}{\bigcup} \mathcal{T}^{t}_\kappa\bigr)$. The first term denotes the number of template pools being updated (using equation (\ref{equ:equ_15})) and the second term denotes the number of template pools being created (using equation (\ref{equ:equ_16})). To reduce the computational burden related to future association, we further restrict the number of templates associated with any template pool to $K$. If the number of associated templates reaches $K+1$, the most irrelevant one is discarded, based on the least likelihood measure given by:
\begin{equation}
n_{d,\kappa}^t = \underset{d \in [1,K+1]}{\mathrm{argmin}}~exp\Bigl(-\frac{1}{2\hat{\xi}} \parallel n_{d,\kappa}^t- S^*_{\textbf{III},\kappa}(:,:,d) \parallel^2 \Bigr) 
\label{equ:equ_17}
\end{equation}
where, $n_{d,\kappa}^t$ is the $d^{th}$ sample which is to be discarded from the updated $\kappa^{th}$ template pool and $S^*_{\textbf{III},\kappa}(:,:,d)$ is the low frequency approximation of the $d^{th}$ template computed from inverse 3D DCT operations. In this way, we conform to the criterion of reliability preservation of a target by keeping only the $K$ most likely appearances of it over a limited duration within a template pool. This step is necessary to dynamically adapt to the changing scene of the video and also to reduce unnecessary computational overhead and accumulated error over time. The overall procedure of our multi-object detection with association step is described in algorithm~\ref{algo:algo1}.

\subsection{Local motion descriptor generation}
We now present a temporal saliency guided optic flow map for accurately describing local motions in our video frames. The details of this formulation is presented below. 
\subsubsection{Generation of MOF}
There exist several methods for determining temporal saliency~\cite{Liu_2014, Gangapure_2018}. We choose the recently proposed method of~\cite{Gangapure_2018} because of its accuracy and execution speed. Let, the OF vector for each pixel $(x,y)$ be denoted as ($\textbf{u(x,y)}$,$\textbf{v(x,y)}$) \cite{Sun_2010} and the corresponding temporal saliency score be denoted as $s(x,y)\in [0,1]$ \cite{Gangapure_2018}. We propose a temporal saliency modulated optic flow vector (MOF), denoted by ($\mathbf{u'(x,y)}$,$\mathbf{v'(x,y)}$) using the modulating function $\mathcal{F}(s)$ as:
\begin{equation}
  \begin{aligned}
   u'(x,y) &= \mathcal{F}(s(x,y)) \cdot u(x,y) \\
   v'(x,y) &= \mathcal{F}(s(x,y)) \cdot v(x,y)
  \end{aligned}
    \label{equ:equ_18}
\end{equation}
The function $\mathcal{F}(s)$ modulates each OF component ($\textbf{u},\textbf{v}$) based on a parameter $\tau \in [0,1]$. For any pixel, if the corresponding saliency value is greater than $\tau$, the flow value is enhanced and if it is less than $\tau$, the flow value is diminished. This is mathematically represented as: 
\begin{equation}
\mathcal{F}\left(s\right)=\ \left\{\begin{array}{l}e^{\frac{s-\tau{}}{2\tau{}}}\ \ \ \ \ \ \ \ \ \ \ \
for\ s\le\tau{} \\
e^{\frac{s}{2}}\ \ \ \ \ \ \ \ \  \ \ \ \ \ for\
s>\tau{}\end{array}\right.
\label{equ:equ_19}
\end{equation}
As the modulating function $\mathcal{F}$ enhances and diminishes the motion magnitude based on the saliency score, we also term it as reward-penalty function. The parameter $\tau$ is determined adaptively at each frame based on the histogram of saliency score for which the area under its cumulative distribution function (CDF) is less than $\alpha$, \textit{i.e.}, $\tau = \underset{\tau'}{argmax}\bigl[ P(\tau' \le s)\le \alpha\bigr]$, where $\tau'$ is a random variable within $[0,1]$. The nature of the function $\mathcal{F}$ for different values of $\tau$ is shown in figure~\ref{fig:figure6}. The range of modulation, as can be seen from equation~(\ref{equ:equ_19}), is restricted to $[e^{-\frac{1}{2}},e^{\frac{1}{2}}]$. This range is suggested keeping in mind the fact that a bigger range would unduly emphasize the enhanced flow values over the diminished ones while a smaller range will result in negligible effect of modulation.
\begin{figure}[h!]
\centering
\includegraphics[width=0.79\textwidth]{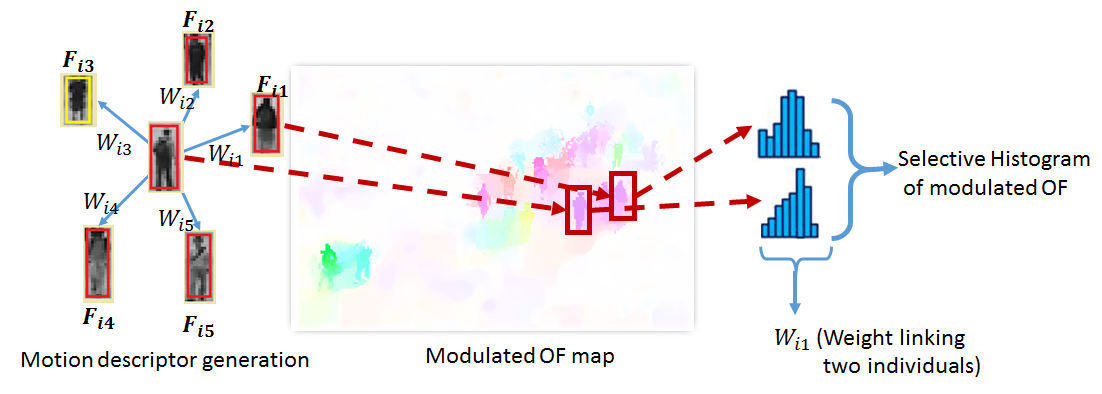}
\caption{A schemetic diagram showing motion descriptor generation using SHMOF computed from MOF map within the region of detected object bounding boxes.}
\label{fig:figure7}
\end{figure}
\subsubsection{Generation of SHMOF}
Let the histogram of MOF (HMOF) maps for the bounding boxes with the $i^{th}$ and $j^{th}$ pedestrians be respectively represented as $H_i$ and $H_j$. These histograms are circularly shifted such that their bin with the highest magnitude becomes centered and are denoted as $H_{i,shifted}$ and $H_{j,shifted}$. Then a selective HMOF (SHMOF) is generated  by limiting the range of HMOF by a parameter $\zeta$ such that the selective-histograms  $H^{\zeta}_{i,shifted}$ and $H^{\zeta}_{j,shifted}$ also contain the bin with maximum magnitude at its center \cite{Yuan_Yuan_2015}. The parameter $\zeta$ can be varied over the range $[0,1]$ such that $\zeta = 0$ represents selection of the bin with only maximum magnitude and $\zeta = 1$ represents the whole HMOF selection. Finally, a bin-to-bin histogram difference $\Delta h^{\zeta}_{i,j}$ is computed between the two SHMOF using the $\chi^2$-distance, given by:
\begin{equation}
\Delta h^{\zeta}_{i,j} = \chi^2(H^{\zeta}_{i,shifted},H^{\zeta}_{j,shifted})
\label{equ:equ_20}
\end{equation} 
where, $\chi^2(H_i,H_j)=\frac{1}{2} \sum^{N_{bins}}_{n=1}{\frac{(H_{i;n}-H_{j;n})^2}{H_{i;n}+H_{j;n}}} $ and $N_{bins}$ denotes the number of bins selected by the parameter $\zeta$. Thus a motion difference vector $\Delta \textbf{h}^{\zeta}_i$ for the $i^{th}$ individual with its $M$ neighbors may be represented as $\Delta \textbf{h}^{\zeta}_i =[\Delta h^{\zeta}_{i,1},\Delta h^{\zeta}_{i,2},...,\Delta h^{\zeta}_{i,M} ]$, where $j = 1,\cdots, M$. 
The value of the parameter $\zeta$ is set according to~\cite{Yuan_Yuan_2015} in our experiments.
\begin{algorithm}[b!]
\caption{Frame Anomaly detection}
\begin{algorithmic}[1]
\Require Video Sequence $\{frame^{t}\},~\beta$
\Ensure  $frame^{t} \leftarrow \Omega,~\overline{\Omega}$
\State \textit{Initilization:} $M \leftarrow 5,~\alpha\leftarrow 70,~\zeta\leftarrow0.9$
\For {t=1 \textbf{to} T} /* $T$ is the total no. of frames */
\State Compute OF map $\{u(x,y),v(s,y)\}$
\State Compute Saliency map $\{S(x,y)\}$
\State Compute MOF map $\{u'(x,y),v'(x,y)\}$ using equation~\ref{equ:equ_18}
\For {i=1 \textbf{to} $\beta$}
\State Determine M nearest neighbor
\State Build motion descriptor $\{W_i,F_i\}^{M}_{k=1}$ from equation~\ref{equ:equ_20} \&~\ref{equ:equ_21}
\State Compute $FA^{t}_{raw}$ over $\beta$ observers
\State Obtain filtered EMD as $FA^{t}$ using equation~\ref{equ:equ_23}
\If {$FA^{t} > threshold$}
\State $frame^{t} \leftarrow \Omega$ 
\Else
\State $frame^{t} \leftarrow \overline{\Omega}$
\EndIf
\EndFor
\EndFor
\end{algorithmic}
\label{algo:algo2}
\end{algorithm}
\subsubsection{Local Descriptor construction} A local motion descriptor is designed from the above motion statistics for each individual over all the $\beta$ observers as obtained in section~\ref{sec:template update} , where, $i = 1,\cdots,\beta$. The local descriptor for the $i^{th}$ individual is represented as $\{\textbf{W}_i,\textbf{F}_i\}^{M}_{k=1}$, where $\mathbf{W}_i \in \mathbb{R}^{1 \times M}$ is the linking weight vector of the $i^{th}$ individual with its $M$ neighbors and $\textbf{F}_i \in \mathbb{R}^{4 \times M}$ is the four-dimensional feature vector describing the connection of each neighbor with the $i^{th}$ individual. The feature vector of each individual consists of 4 elements, namely, maximum, minimum, mean and variance of motion energy of the individual regions of MOF map enclosed by their corresponding bounding boxes. The weight term of the motion descriptor links an individual to its neighbors based on the motion difference vectors. The normalized linking weight $W_{i,k}$ between the $i^{th}$ individual and its $k^{th}$ neighbor is given by:
\begin{equation}
W_{i,k} = \frac{(\Delta h^{\zeta}_{i,k})^2}{\sum^{M}_{k=1} (\Delta h^{\zeta}_{i,k})^2}
\label{equ:equ_21}
\end{equation}
Based on equation~(\ref{equ:equ_21}) each component of the motion difference vector $\Delta \textbf{h}_i^{\zeta}$ is scaled over the quadratic function to give significant weight to large motion changes. Thus, the more distinct the motion of an individual with respect to its surrounding, the larger will be its linking weight. The feature vectors along with their linking weights form a distinctive signature between an observer and its surrounding individuals.  
\subsection{Crowd Anomaly analysis}
\label{section D}
Anomaly in a frame is determined by an overall abrupt change in motion. This abrupt change in motion can be measured by comparing local motion between consecutive frames using suitable motion descriptors. Our motion descriptors are generated locally to represent complex motion dynamics.  Cumulative local changes of the descriptors are measured using EMD. If the magnitude of EMD exceeds a certain experimentally chosen threshold, the frame is deemed as anomalous.  
\begin{figure*}[t!]
\centering  
\subfigure[]{\includegraphics[width=0.32\linewidth]{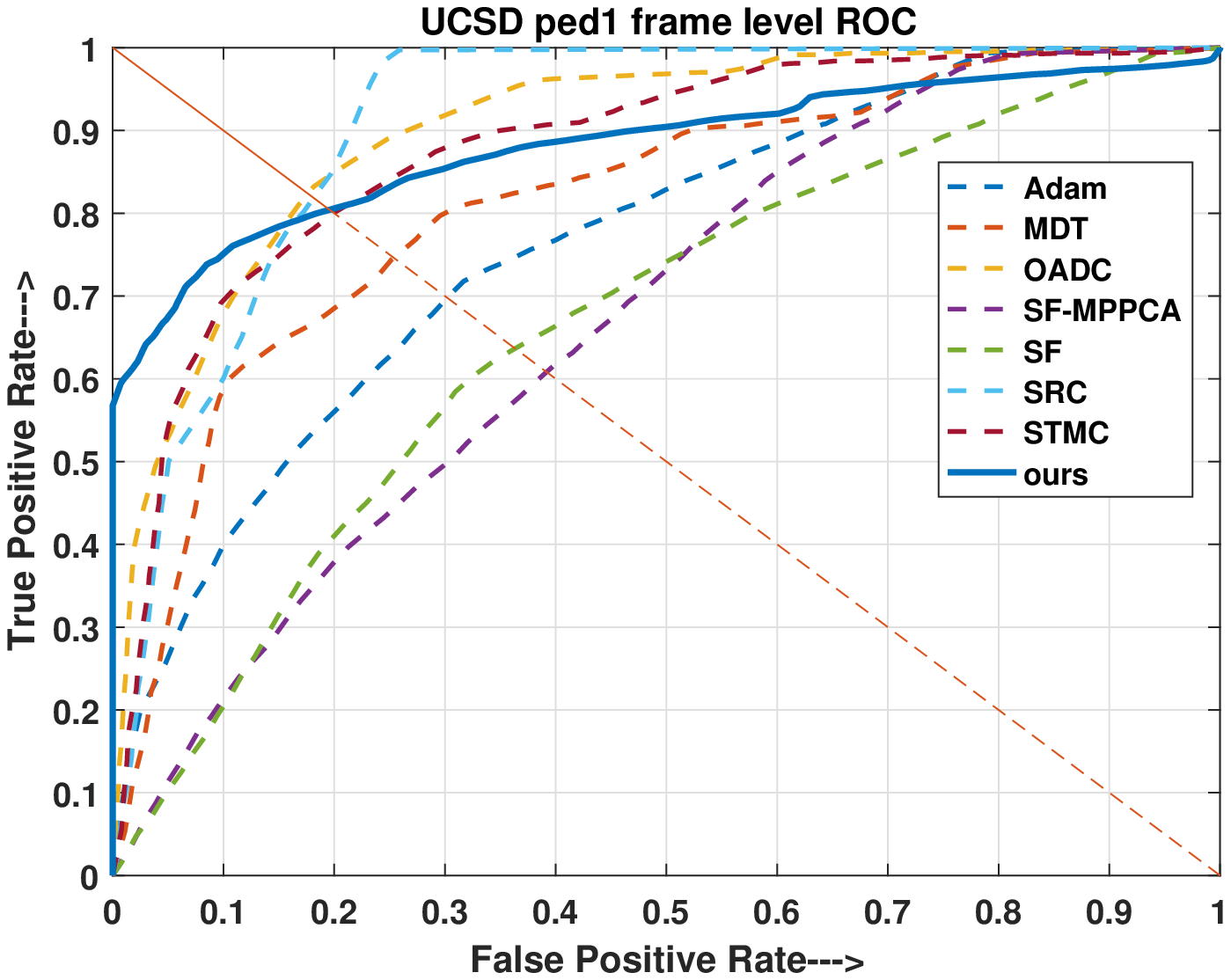}}
\subfigure[]{\includegraphics[width=0.32\linewidth]{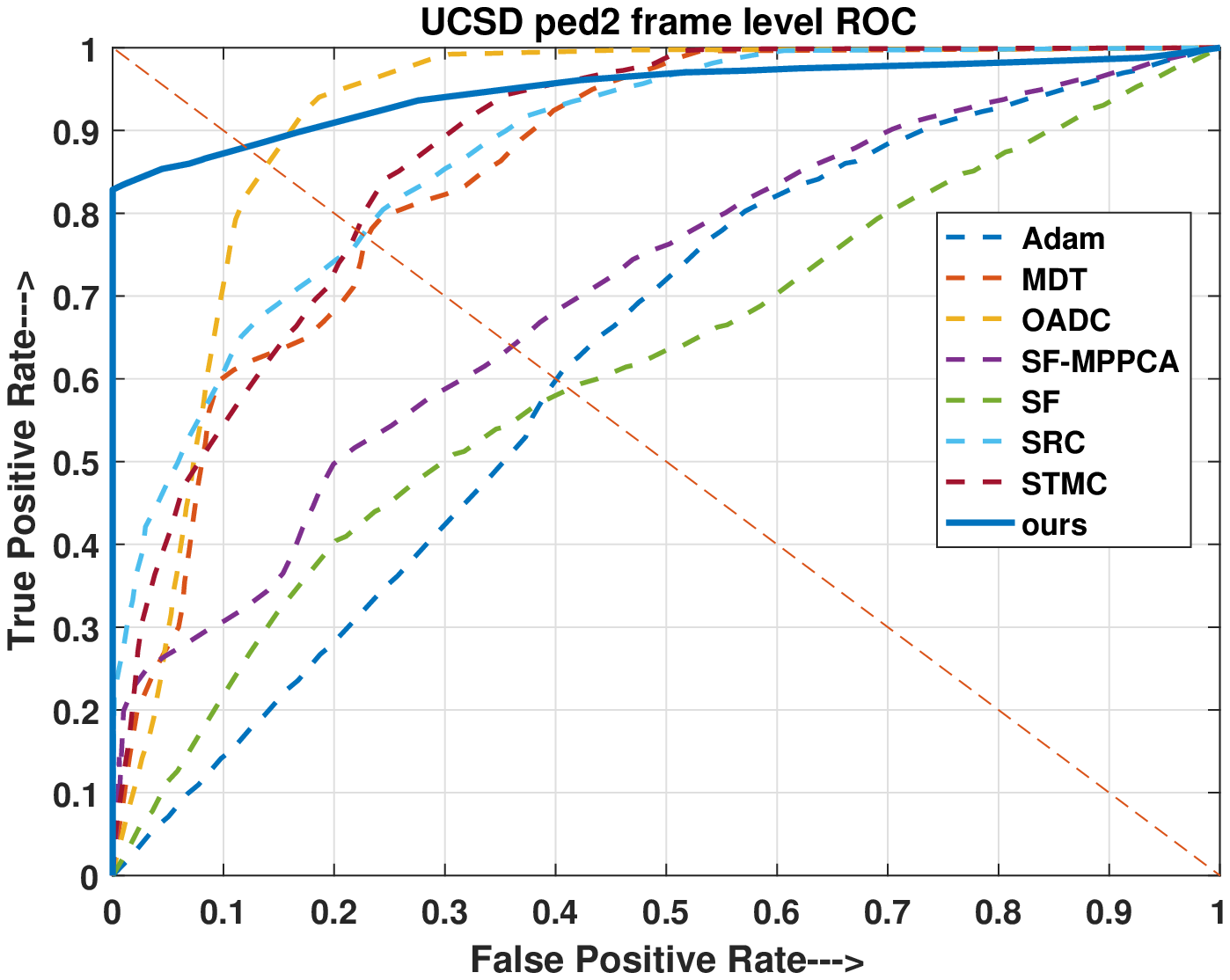}}
\subfigure[]{\includegraphics[width=0.32\linewidth]{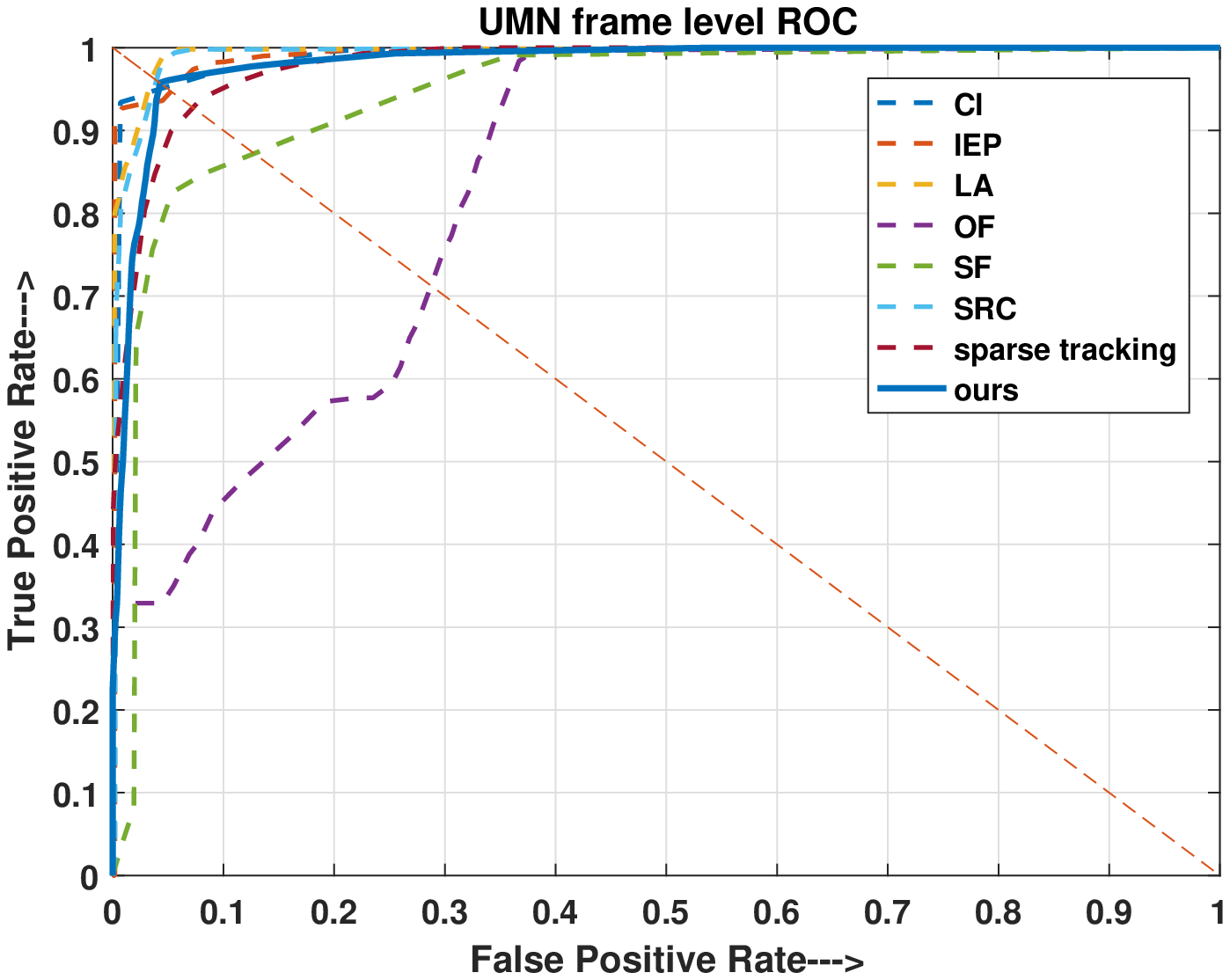}}
\caption{Comparisons of Frame-level ROC over dataset UCSD \textit{ped1}, UCSD \textit{ped2} dataset and UMN video sequences in figures (a), (b) and (c) respectively.} 
\label{fig:figure10}
\end{figure*}    
Frame-level abnormality is detected by comparing corresponding $\beta$ observers over two consecutive frames based on the local structures captured by the descriptor $\{\textbf{W}_i,\textbf{F}_i\}^{M}_{k=1}$. A relative variation between these local structures are captured between the $i^{th}$ observer at frame $(t-1)$, \textit{i.e.}, $\{\textbf{W}_i^{t-1},\textbf{F}_i^{t-1}\}^{M}_{k=1}$ and at frame $t$, \textit{i.e.}, $\{\textbf{W}_i^{t},\textbf{F}_i^{t}\}^{M}_{k=1}$ using EMD~\cite{Rubner_2001} and is expressed as:
\begin{equation}
FA_{raw}^{t} = \frac{1}{\beta} \sum^{\beta}_{i=1}{EMD(\textbf{W}_i^{t-1},\textbf{W}_i^{t},\textbf{F}_i^{t-1},\textbf{F}_i^{t})}
\label{equ:equ_22}
\end{equation}
where, $FA_{raw}^{t}$, computed for the $t^{th}$ frame, is termed as raw abnormality value and is often stochastic in nature. Sometimes, a high $FA_{raw}^{t}$ value can result from erroneous detection and can also render a normal frame abnormal. Therefore, a post-filtering operation is performed to smooth out this effect using a weighted mean filter. The new measure is defined as:
\begin{equation}
FA^t = \frac{1}{2n+1} \sum^{t+n}_{t' = (t-n)} wt * FA_{raw}^{t'}
\label{equ:equ_23}
\end{equation}
where, $wt$ is a partial weight factor given by
\begin{equation}
wt = \left\{ \begin{array}{c}
\frac{1}{n+1}~~~~~\ \ \ \ \ \ \ if\ \ \ t' = n \\ 
\frac{1}{2(n+1)}~~~~~\ \ \ \ \ \ if\ \ \ \ t' \neq n \end{array}
\right.
\label{equ:equ_24}
\end{equation}
The filter length is $(2n+1)$, where, $n$ is an integer. We have used $n = 3$ in this paper. A frame is deemed anomalous ($\Omega$) if its $FA^{t} > 0.5$ and normal ($\overline{\Omega}$) otherwise. The threshold value of $0.5$ is set according to \cite{Yuan_Yuan_2015}.
\begin{table*}[h]
  \centering
  \caption{Performance measure showing AUC values on frame-level anomaly for different frameworks on UMN dataset}
  \scalebox{0.45}{ 
    \begin{tabular}{rrccccccccrrrr}
          &       &       &       &       &       &       &       &       &       &       &       &       &  \\
\cmidrule{2-13}    \multicolumn{1}{r|}{} & \multicolumn{1}{c|}{\textbf{Methods}} & \multicolumn{1}{c|}{\textbf{OF}~\cite{Mehran_2009}} & \multicolumn{1}{c|}{\textbf{SF}~\cite{Mehran_2009}} & \multicolumn{1}{c|}{\textbf{CI}~\cite{Wu_2010}} & \multicolumn{1}{c|}{\textbf{IEP}~\cite{Cui_2011}} & \multicolumn{1}{c|}{\textbf{LA}~\cite{Saligrama_2012}} & \multicolumn{1}{c|}{\textbf{SRC}~\cite{Cong_2013}} & \multicolumn{1}{c|}{\textbf{Matrix approx.}~\cite{Wang_2012}} & \multicolumn{1}{c|}{\textbf{Tracking sparse comp.}~\cite{Biswas_2016}} & \multicolumn{1}{c|}{\textbf{H-MDT-CRF}~\cite{Weixin_Li_2014}} & \multicolumn{1}{c|}{\textbf{OADC-SC}~\cite{Yuan_Yuan_2015}} & \multicolumn{1}{c|}{\textbf{Ours}} &  \\
\cmidrule{2-13}    \multicolumn{1}{r|}{} & \multicolumn{1}{c|}{Scene1} & \multicolumn{1}{c|}{$-$} & \multicolumn{1}{c|}{$-$} & \multicolumn{1}{c|}{$-$} & \multicolumn{1}{c|}{$-$} & \multicolumn{1}{c|}{$-$} & \multicolumn{1}{c|}{0.995} & \multicolumn{1}{c|}{0.994} & \multicolumn{1}{c|}{0.9997} & \multicolumn{1}{c|}{$-$} & \multicolumn{1}{c|}{$-$} & \multicolumn{1}{r|}{\textbf{0.988}} &  \\
\cmidrule{2-13}    \multicolumn{1}{r|}{} & \multicolumn{1}{c|}{Scene2} & \multicolumn{1}{c|}{$-$} & \multicolumn{1}{c|}{$-$} & \multicolumn{1}{c|}{$-$} & \multicolumn{1}{c|}{$-$} & \multicolumn{1}{c|}{$-$} & \multicolumn{1}{c|}{0.975} & \multicolumn{1}{c|}{0.971} & \multicolumn{1}{c|}{0.931} & \multicolumn{1}{c|}{$-$} & \multicolumn{1}{c|}{$-$} & \multicolumn{1}{r|}{\textbf{0.978}} &  \\
\cmidrule{2-13}    \multicolumn{1}{r|}{} & \multicolumn{1}{c|}{Scene3} & \multicolumn{1}{c|}{$-$} & \multicolumn{1}{c|}{$-$} & \multicolumn{1}{c|}{$-$} & \multicolumn{1}{c|}{$-$} & \multicolumn{1}{c|}{$-$} & \multicolumn{1}{c|}{0.964} & \multicolumn{1}{c|}{0.996} & \multicolumn{1}{c|}{0.834} & \multicolumn{1}{c|}{$-$} & \multicolumn{1}{c|}{$-$} & \multicolumn{1}{r|}{\textbf{0.984}} &  \\
\cmidrule{2-13}    \multicolumn{1}{r|}{} & \multicolumn{1}{c|}{Overall} & \multicolumn{1}{c|}{0.84} & \multicolumn{1}{c|}{0.96} & \multicolumn{1}{c|}{0.99} & \multicolumn{1}{c|}{0.989} & \multicolumn{1}{c|}{0.985} & \multicolumn{1}{c|}{$-$} & \multicolumn{1}{c|}{$-$} & \multicolumn{1}{c|}{$-$} & \multicolumn{1}{c|}{0.955} & \multicolumn{1}{c|}{0.9967} & \multicolumn{1}{c|}{\textbf{0.982}} &  \\
\cmidrule{2-13}          &       &       &       &       &       &       &       &       &       &       &       &       &  \\
    \end{tabular}}%
  \label{tab:tab2}%
\end{table*}%

Unlike other forms of anomalies in video scenes~\cite{Li_2015,Sodemann_2012}, crowd anomaly mostly encounters complex scene dynamics. Non-stationarity of anomaly in both spatial and temporal regions are one of the key concerns addressed in this work. To handle such problems, we generate reliable motion descriptors via continually destroying and creating new template pools over stable observers (section~\ref{sec:template update}). Moreover, as a stable object (observer) shifts spatio-temporally over the frame, the descriptors are generated in corresponding regions. This is illustrated in figure~\ref{fig:figure18}. Thus our model automatically handles non-stationarity in crowd anomaly.
\section{Experimental Results}
\label{sec_experiments}
\subsection{Dataset Description}
We have experimented with four publicly available datasets, namely, UCSD, UMN, CHUK-Avenue and ShanghaiTech. UCSD datset~\cite{Li_2013} is mostly adopted for crowd anomaly. Common anomalies include entering of undesirable objects like bikers, skaters, small carts into a pedestrian walkway. Other instances of anomalies include unusual variation of crowd density. Unlike UCSD, UMN dataset~\cite{Yuan_Yuan_2015} only consists of persons and anomaly described in this dataset points to a crowd escape behavior. CUHK Avenue~\cite{Lu_2013} is another dataset with abnormal events such as loitering and running, throwing objects. We have also evaluated our model over a recently proposed Shanghai Campus dataset~\cite{Luo_2017}. It consists of many scenes with complex viewing angles and lightning conditions.  
 

\begin{figure}[h!]
\centering
\includegraphics[width=0.69\textwidth]{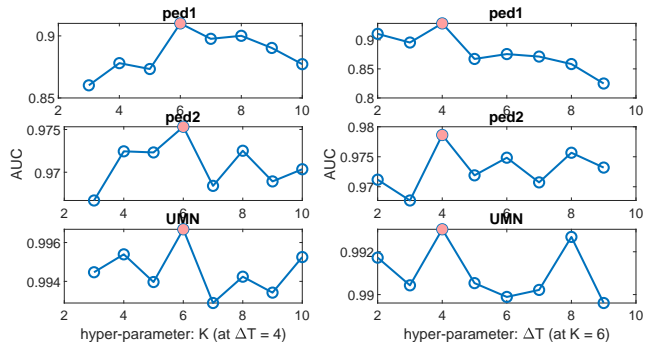}
\caption{Effect of different hyper-parameters on AUC over UCSD ped1, UCSD ped2 and UMN dataset.}
\label{fig:figure17}
\end{figure}
\begin{figure}
\centering
\includegraphics[width=0.62\textwidth]{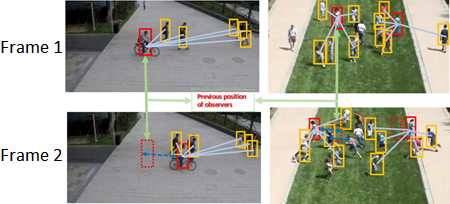}
\caption{The figure shows the non-stationarity of local descriptors where observers with their surrounding individuals are represented by red and yellow bounding boxes respectively. The dotted bounding boxes in frame 2 show the previous position of the observers.}
\label{fig:figure18}
\end{figure}
\begin{table}[t!]
  \centering
  \caption{Performance measure showing AUC on frame-level anomaly for different frameworks over UCSD dataset}
  \scalebox{0.7}{
    \begin{tabular}{|c|c|c|c|c|}
    \toprule
    \multirow{3}[6]{*}{Method} & \multicolumn{4}{c|}{Criterion} \\
\cmidrule{2-5}          & \multicolumn{2}{c|}{AUC} & \multicolumn{2}{c|}{EER} \\
\cmidrule{2-5}          & Ped1  & Ped2  & ped1  & ped2 \\
    \midrule
    Adam~\cite{Adam_2008}  & 0.65  & 0.63  & 38\%  & 42\% \\
    \midrule
    MPPCA~\cite{Kim_Grauman_2009} & 0.59  & 0.71  & 32\%  & 36\% \\
    \midrule
    SF~\cite{Mehran_2009}    & 0.67  & 0.63  & 31\%  & 42\% \\
    \midrule
    MDT~\cite{Mahadevan_2010}   & 0.818 & 0.85  & 25\%  & 25\% \\
    \midrule
    SRC~\cite{Cong_2013}   & 0.86  & 0.861 & \_    & \_ \\
    \midrule
    Matrix approx.~\cite{Wang_2012}   & 0.772  & 0.931 & 31\%   & 15\% \\
    \midrule
   Tracking Sparse comp.~\cite{Biswas_2016}   & 0.887  & 0.967 & 18\%   & 8\% \\
    \midrule
    STMC~\cite{Yang_Cong_2013}  & 0.88  & 0.868 & \_    & \_ \\
    \midrule
    OADC-SA~\cite{Yuan_Yuan_2015} & 0.91  & 0.925 & \_    & \_ \\
    \midrule
    \textbf{Ours}  &  \textbf{0.882}     & \textbf{0.942} &   \textbf{22\%}    & \textbf{13\%} \\
    \bottomrule
    \end{tabular}}%
  \label{tab:tab1}%
\end{table}%
\subsection{Hyper-parameter Settings}
In this section, we provide the details of hyper-parameter settings. Since, we are mostly dealing with pedestrians in crowd scenes, the aspect ratio of the bounding boxes are kept within a range of $[0.33,0.49]$. This range is chosen typically in accordance to the pedestrian detection work of Dollar \textit{et al.}~\cite{Dollar_2012} where they have stated that the general aspect ratio of a person is $\sim 0.41$. The threshold parameters $\phi_{thres}$ and $\psi_{thres}$ (as described in section~\ref{lab:bbox_generation}) for generating appropriate proposal bounding boxes are experimentally set to $0.5$ and $0.7$ respectively. This is described in figure~\ref{fig:figure9}. This setting works for most of the video scenes. We have also analyzed variations of AUC over other two hyper-parameters $\Delta T$ and $K$ following figure~\ref{fig:figure17}. By analyzing the curves we empirically set $K$ = 6 and $\Delta T$ = 4.
\begin{figure}[th!]
\centering
    \centering
    \begin{subfigure}
        \centering       
$\begin{array}{cccc}
\includegraphics[width=0.35\textwidth]{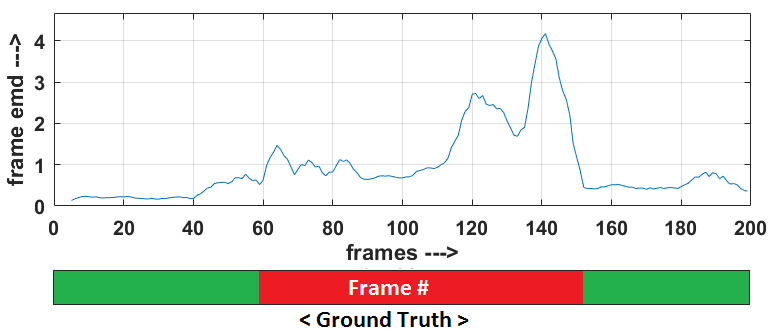}&
 \includegraphics[width=0.35\textwidth]{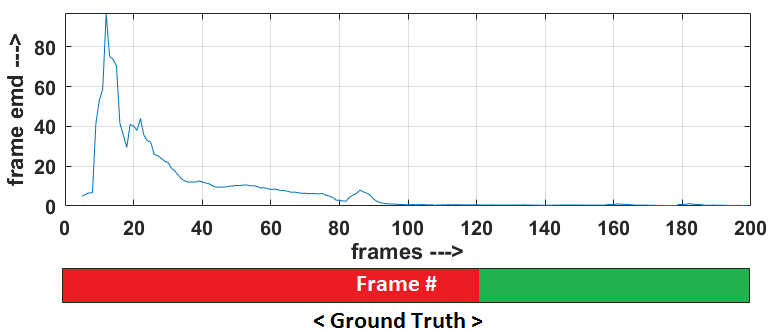} \\
 \includegraphics[width=0.35\textwidth]{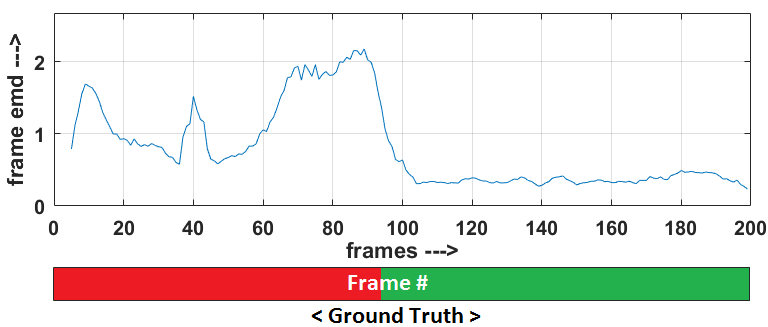}  &
 \includegraphics[width=0.35\textwidth]{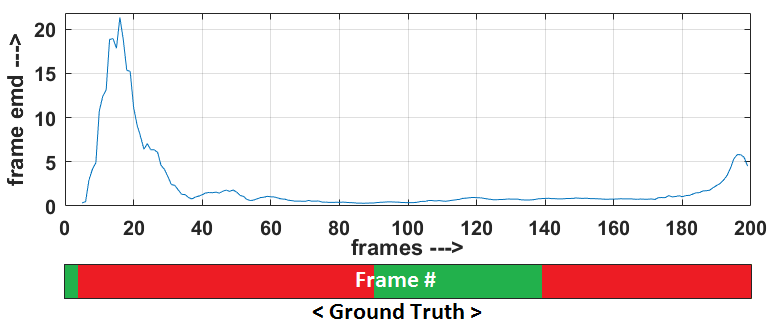}
\end{array}$\\
\textit{UCSD ped1 dataset}\\
    \end{subfigure}%
    ~
        \begin{subfigure}
        \centering       
$\begin{array}{cccc}
\includegraphics[width=0.35\textwidth]{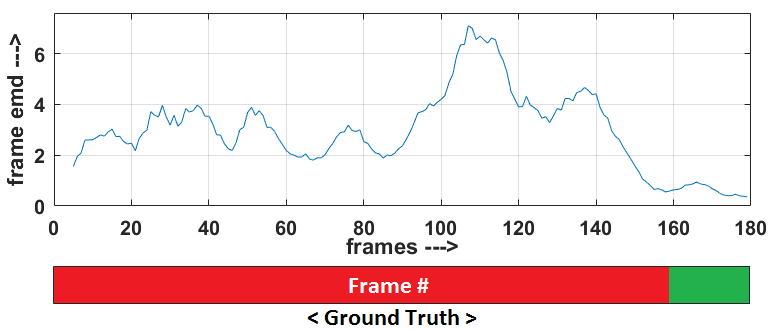}&
 \includegraphics[width=0.35\textwidth]{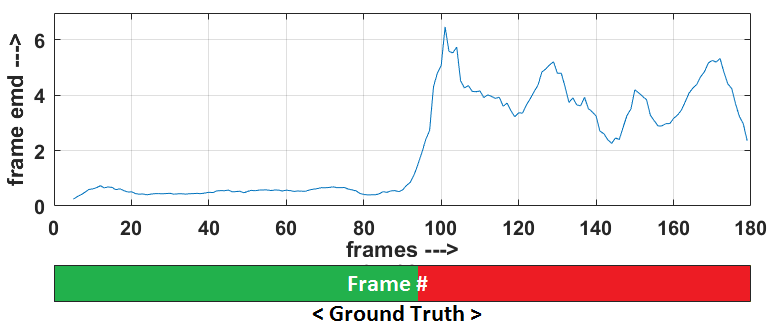} \\
 \includegraphics[width=0.35\textwidth]{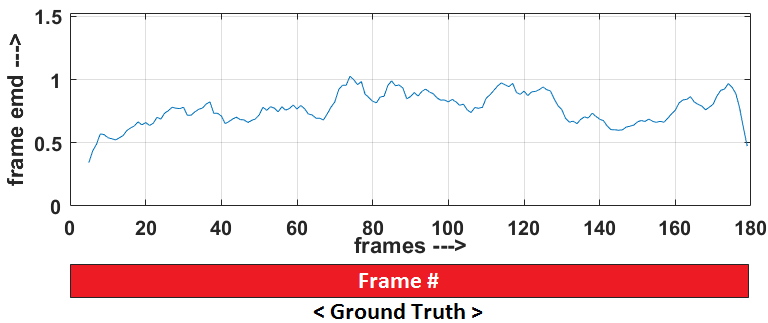}  &
 \includegraphics[width=0.35\textwidth]{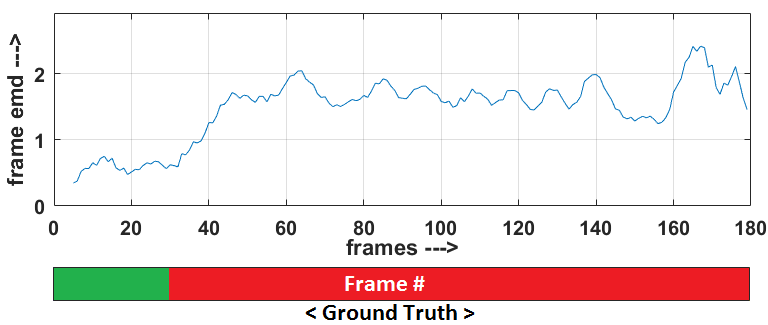}
\end{array}$\\
\textit{UCSD ped2 dataset}\\
    \end{subfigure}%
    ~
        \begin{subfigure}
        \centering       
$\begin{array}{cccc}
\includegraphics[width=0.35\textwidth]{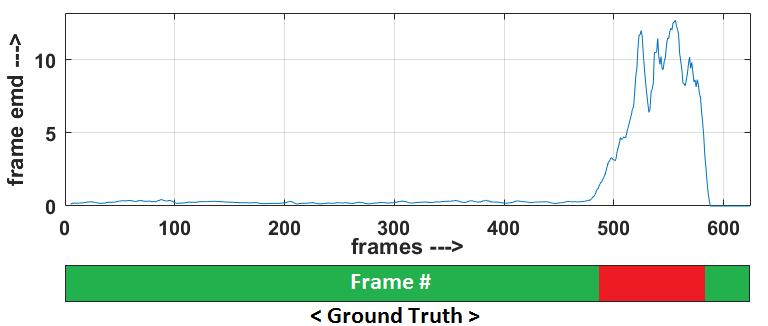}&
 \includegraphics[width=0.35\textwidth]{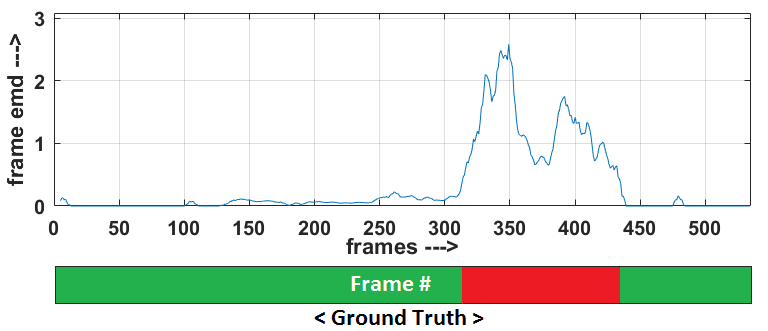} \\
 \includegraphics[width=0.35\textwidth]{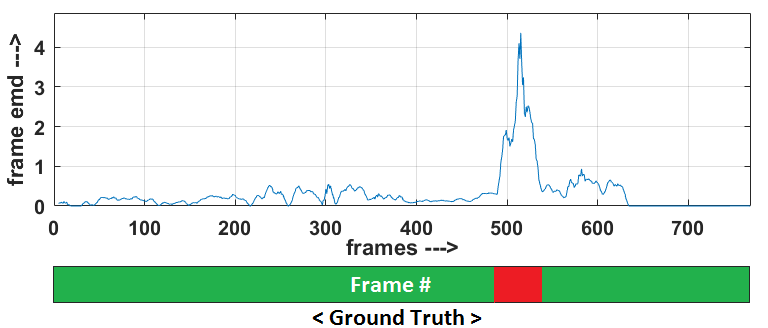}  &
 \includegraphics[width=0.35\textwidth]{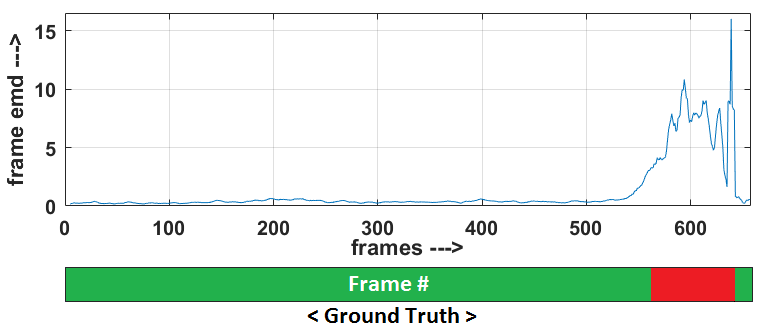}
\end{array}$\\
\textit{UMN dataset}
    \end{subfigure}%
\caption{Frame level EMD variation of several datasets with frame level ground truth annotations (shown in bar below the plots), where green and red bar indicates the normal and abnormal frames respectively. Top two rows: UCSD \textit{ped1} video sequences, Middle two rows: UCSD \textit{ped2} video sequences, Bottom two rows: UMN video sequences} 
\label{fig:figure11}
\end{figure} 
\subsection{Performance Measure}
We have evaluated the efficiency of our proposed pipeline in qualitative as well as in quantitative manner. As a part of the available ground-truth, a frame has been marked either \textquotedblleft anomalous \textquotedblright{} or \textquotedblleft normal \textquotedblright{}. For qualitative evaluation, we plot frame level EMD values vs. ground-truth. For an anomalous frame, the EMD value should be high and for a normal frame, this value should be low. For quantitative evaluation, an anomalous frame is treated as \textquotedblleft positive \textquotedblright{} and a normal frame is treated as \textquotedblleft negative\textquotedblright{}. We use receiver operating characteristics (ROC) curves, area under ROC curves (AUC) and equal error rate (EER) as the three performance measures. ROC denotes the variation of True positive rate (TPR) versus False positive rate (FPR). EER summarizes the ratio of misclassified frames at which $FPR = 1-TRP$. Note that here we have only computed anomaly at frame level. Given the stochastic nature of the descriptor generation and dynamic nature of the scene,measuring pixel level anomaly in a training-less setting like ours becomes extremely cumbersome. 
\subsection{Ablation Study}
\subsubsection{Influence of modulation} 
\begin{figure}
\begin{center}$
\begin{array}{ccc}
\includegraphics[width=0.32\textwidth]{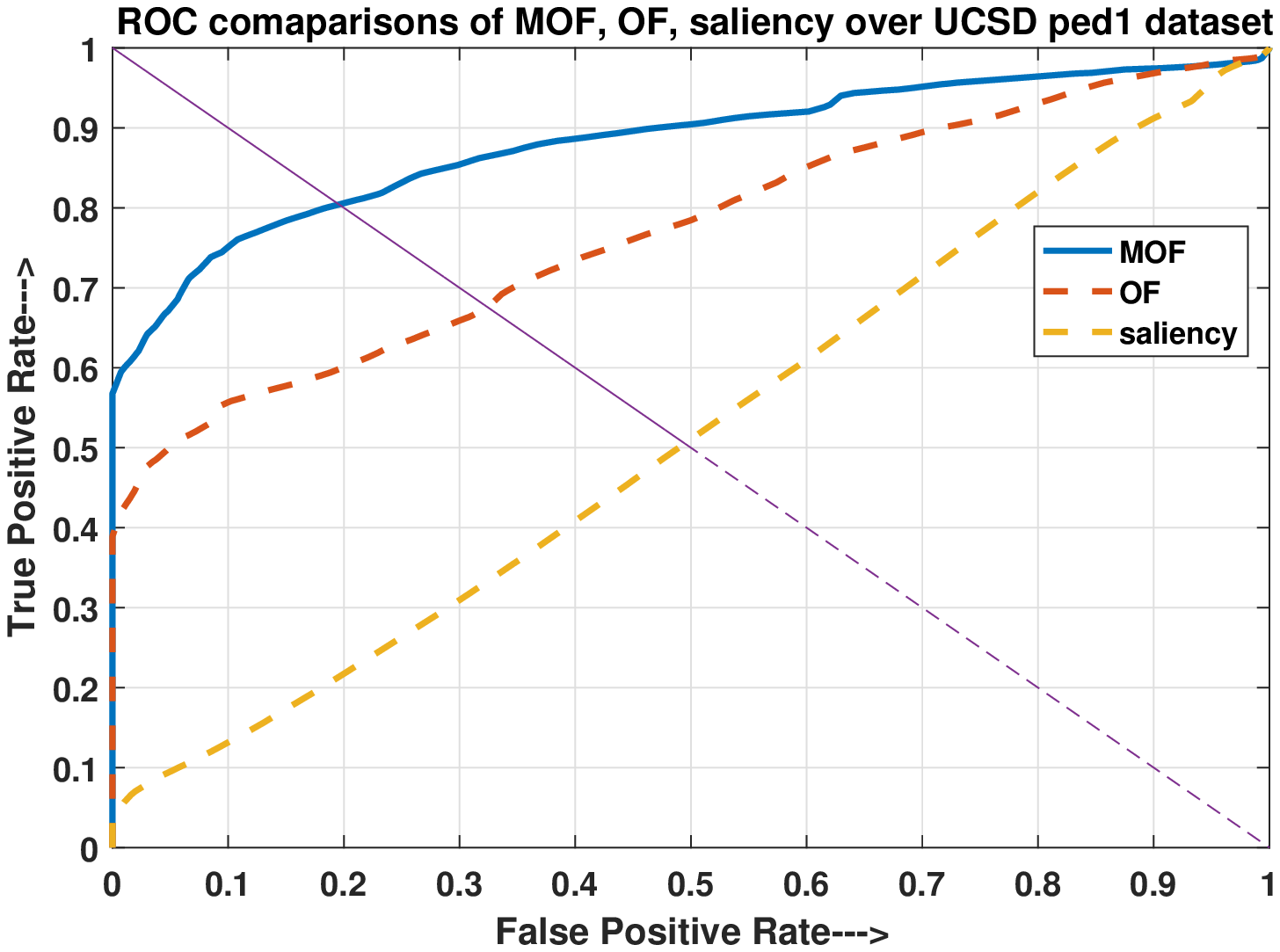}&
 \includegraphics[width=0.32\textwidth]{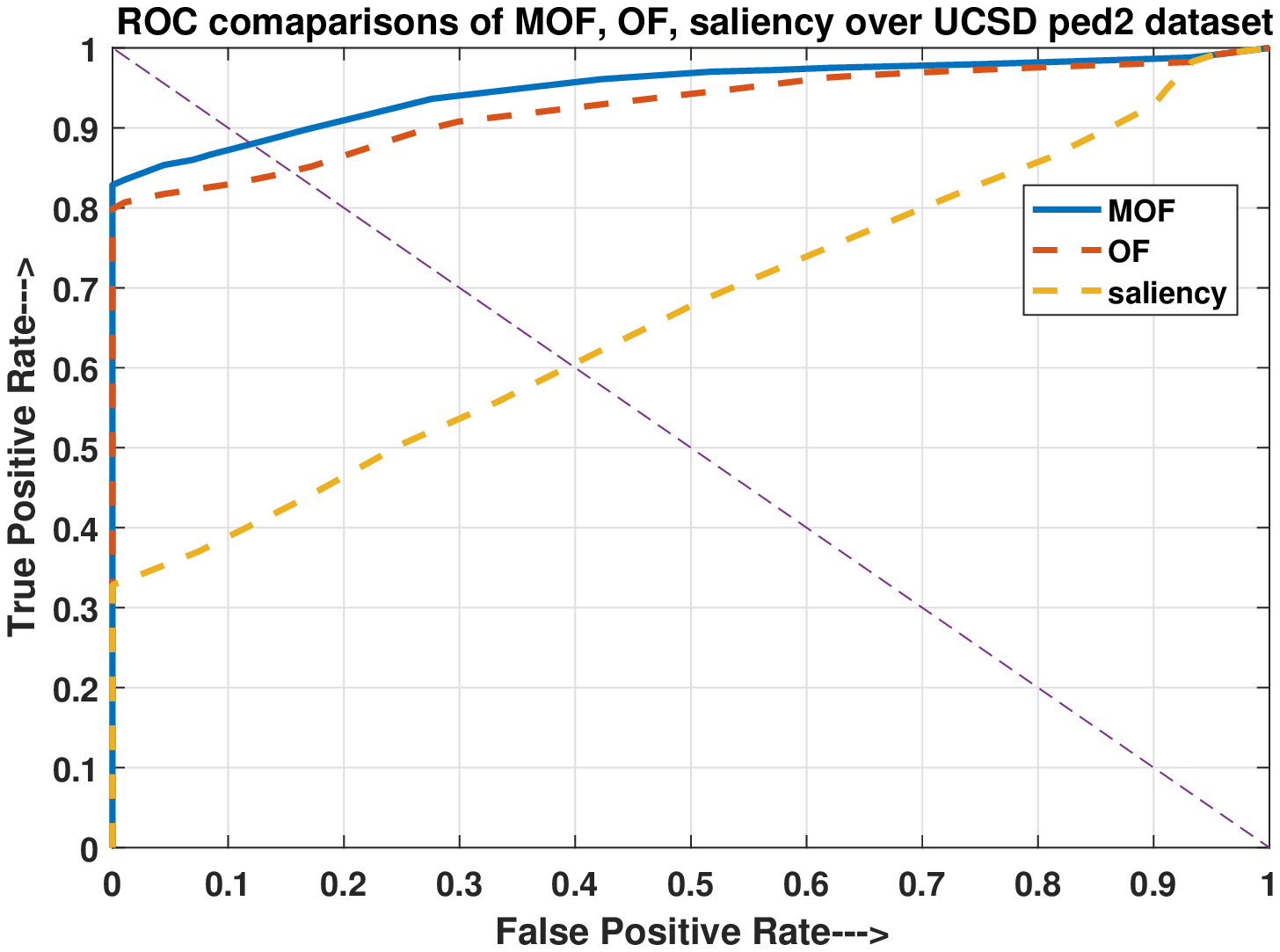} &
 \includegraphics[width=0.32\textwidth]{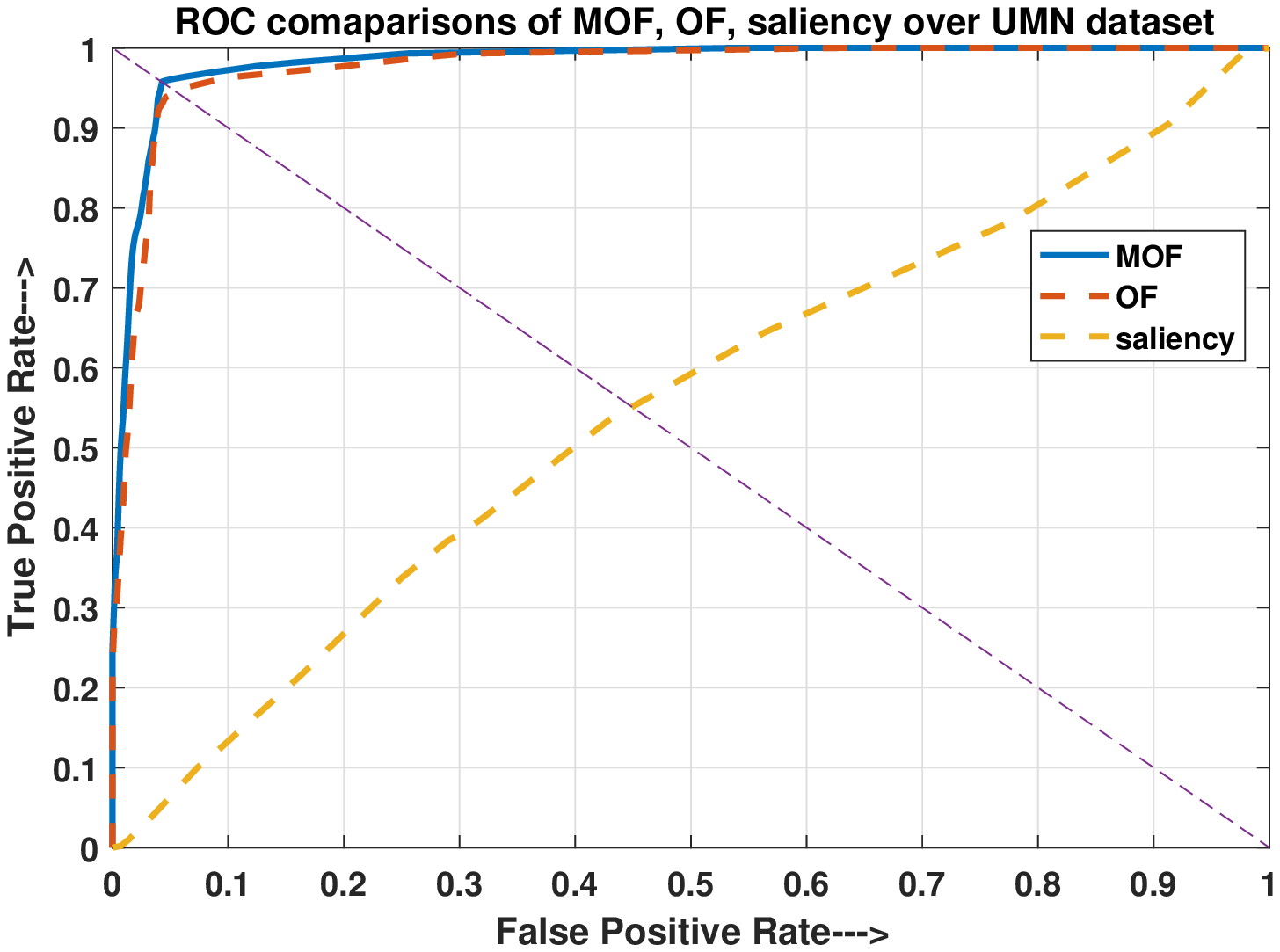}
\end{array}$
\end{center}
\centering
\caption{Comparisons of ROC curves for frame-level criteria for various datasets showing improved performance using MOF over OF and saliency map.} 
\label{fig:figure14}
\end{figure}

\begin{figure}
\centering
\includegraphics[width=.6\textwidth]{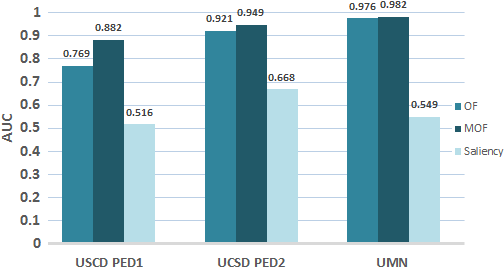}
\caption{Comparison of AUCs using OF, MOF and saliency maps over varying dataset.}
\label{fig:figure15}
\end{figure}

To show the effectiveness of saliency modulated optic flow (MOF), we build ROC curves and compute AUC values for only optic flow (OF), MOF and only saliency on the UCSD \textit{ped1}, UCSD \textit{ped2} and UMN datasets. Please see figure~\ref{fig:figure14} and figure~\ref{fig:figure15} for the ROC curves and AUC plots respectively. It can be seen that the influence of modulation is much more pronounced with an overall AUC of 0.882 from MOF as compared to that of OF (AUC = 0.769) and only saliency (AUC = 0.516) on the more challenging UCSD \textit{ped1} dataset. On UCSD \textit{ped2} and UMN datasets, the improvement of MOF over OF is less pronounced. We have also performed a paired t-test over AUC values of MOF-OF and MOF-saliency pair over all the video sequences. The corresponding p-values, shown in table~\ref{tab:tab3} indicate that the improvements of MOF over that of OF and saliency for all three datasets are statistically significant. 
\begin{table}
\centering
\caption{Table showing improved performance of 3D-DCT based association in terms of AUC using adaptive $\xi$ over fixed $\xi$}
\label{my-label}
  \scalebox{0.9}{
\begin{tabular}{|c|l|l|l|l}
\cline{1-4}
\multirow{2}{*}{Method} & \multicolumn{3}{c|}{AUC}      &  \\ \cline{2-4}
                        & USCD ped1 & UCSD ped2 & UMN   &  \\ \cline{1-4}
fixed $\xi$~\cite{Yuan_Yuan_2015}                     & 0.761     & 0.856     & 0.886 &  \\ \cline{1-4}
Adaptive $\xi$            & 0.882     & 0.942     & 0.982 &  \\ \cline{1-4}
\end{tabular}}
\label{tab:tab4}%
\end{table}
\begin{figure*}[t!]
\centering  
\subfigure[]{\includegraphics[width=0.32\linewidth]{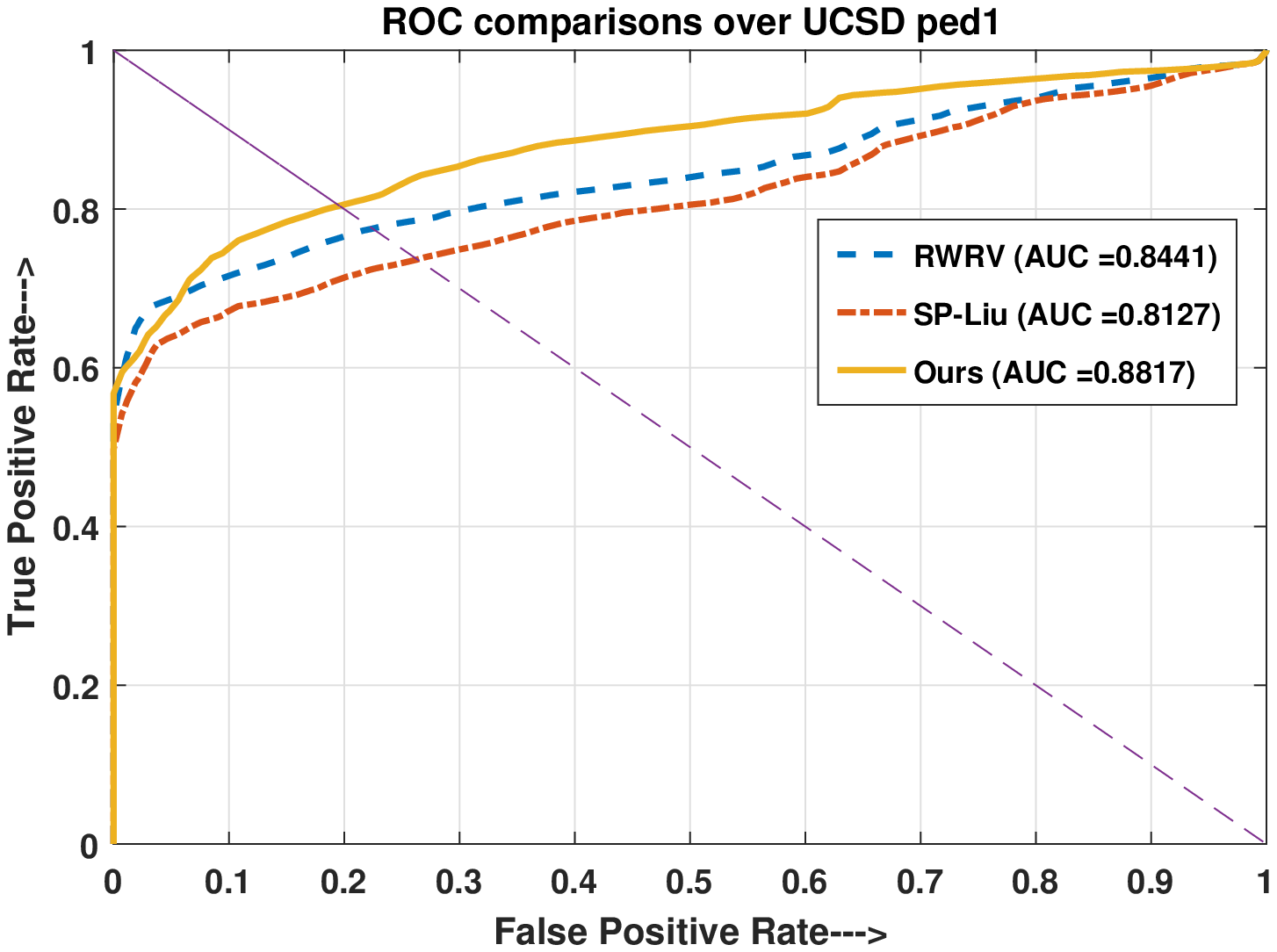}}
\subfigure[]{\includegraphics[width=0.32\linewidth]{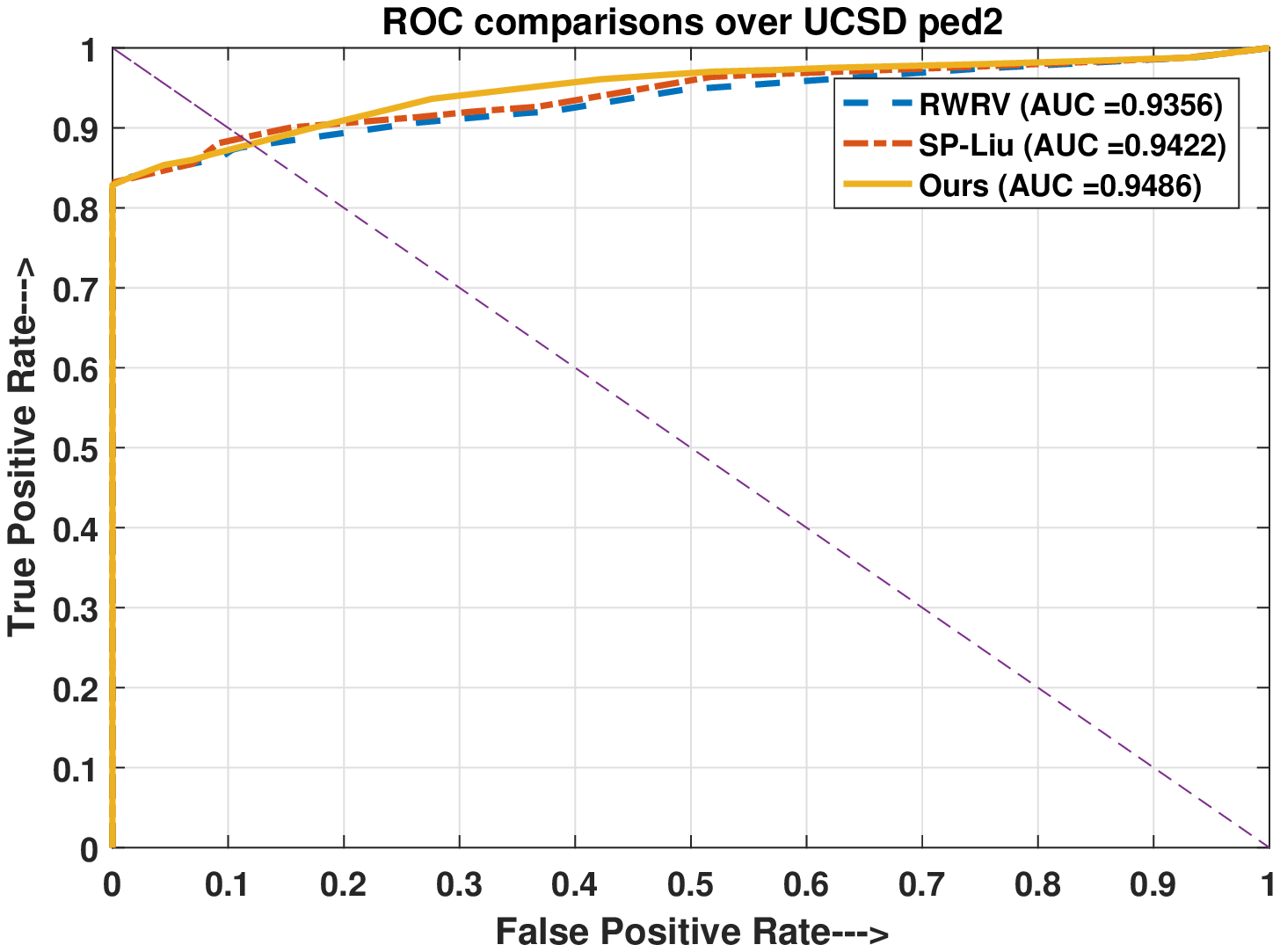}}
\subfigure[]{\includegraphics[width=0.32\linewidth]{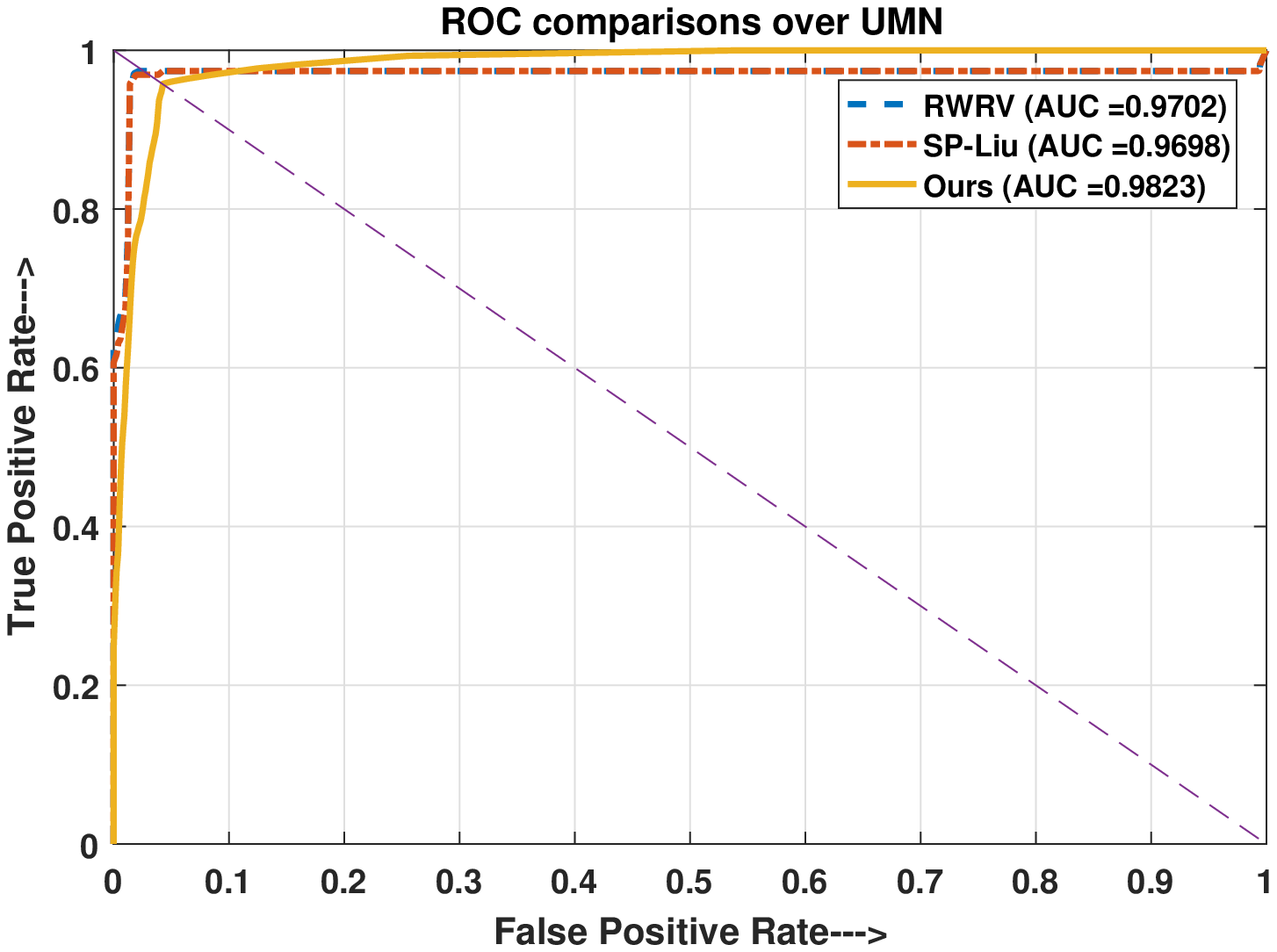}}
\caption{Comparative evaluation and effect of other temporal saliency maps over ROC and AUC for video sequences (a) UCSD \textit{ped1}, (b) UCSD \textit{ped2} and (c) UMN datasets.} 
\label{fig:figure16}
\end{figure*}
\subsubsection{Impact of adaptive 3D-DCT based association} A 3D-DCT based approach is developed for efficiently detecting and associating multiple objects. In Table~\ref{tab:tab4}, we show an improved performance in terms of AUC for adaptive $\xi$ over fixed $\xi$~\cite{Yuan_Yuan_2015}. From the table it can be clearly seen that there is an approximate 10\% increase in the AUC values for both UCSD and UMN datasets which indicate the robustness of our method. We have also demonstrated the improvements resulting from the adaptive 3D-DCT model over the fixed 3D-DCT model in table \ref{tab:tab5}. We take one particular scene from each dataset and computed the variance of likelihood values at every $5^{th}$ and $10^{th}$ frame (frame numbers are chosen without any loss of generality). We have compared the following two cases. In one case, we have kept the scaling parameter  $\xi$ fixed at 0.1 as suggested by OADC-SA. In the second case, we have estimated and set  $\xi$ based on the equation~(\ref{equ:equ_12a}). It can be clearly seen that the variance of the likelihood values is significantly higher for the estimated $\xi$ in all cases. 
\subsection{Results on UCSD datasets}
The methods which have been tested on the UCSD dataset include the one by Adam \textit{et al.}~\cite{Adam_2008}, MPPCA~\cite{Kim_Grauman_2009}, the social force (SF) model~\cite{Mehran_2009}, mixture of dynamic texture (MDT)~\cite{Mahadevan_2010}, sparse reconstruction cost (SRC)~\cite{Cong_2013}, spatio-temporal motion context (STMC)~\cite{Yang_Cong_2013}, anomaly detection method exploiting the crowd structure (OADC-SA)~\cite{Yuan_Yuan_2015} and crowd abnormality detection by tracking sparse components~\cite{Biswas_2016}. For the UCSD \textit{ped1} and \textit{ped2} datasets, ROC curves for various competing methods are presented in figure~\ref{fig:figure10}. Corresponding AUC and EER values are reported in table~\ref{tab:tab1}. The results clearly indicate that the proposed training-less method has attained very competitive results. Our method has achieved an overall AUC of 0.882 on \textit{ped1} dataset which is higher than the AUC values of seven out of the nine competing methods. On \textit{ped2} dataset, we obtain an overall AUC of 0.942 and beat eight out of nine methods. Our EER values of 0.22 and 0.13 in \textit{ped1} and \textit{ped2} dataset respectively are also lower than all but ~\cite{Biswas_2016}. Some qualitative response of our model on video clips of \textit{ped1} and \textit{ped2} are shown in figure~\ref{fig:figure11}(a) and figure~\ref{fig:figure11}(b) respectively. In these figures, we show the variations of frame level EMD with the ground truths. The variations show a hike in response mostly at the anomalous regions.\\ \indent Additionally, we have also performed some comparative study with temporal saliency maps of other methods, namely, SP-Liu~\cite{Liu_2014} and RWRV~\cite{Kim_2013}. Modulation with the temporal saliency map~\cite{Gangapure_2018} (which we have adopted in our method) have attained better performance in terms of AUC for both ped1 (AUC = 0.8817) and ped2 datasets (AUC = 0.9486). Liu \textit{et al.}~\cite{Liu_2014} have achieved an AUC of 0.8122 and 0.9422 and Kim \textit{et al.}~\cite{Kim_2013} have achieved an AUC of 0.8441 and 0.9356 in ped1 and ped2 respectively. We have shown performance comparisons in terms of ROC curves for all the methods in figure~\ref{fig:figure16}.
\begin{table}
\centering
\caption{Showing the variance of likelihood value for every dataset over fixed and adaptive 3D-DCT model}
\scalebox{0.76}{
\begin{tabular}{|c|c|c|c|c|}
\hline
    \multirow{4}[6]{*}{Dataset} & \multicolumn{4}{c|}{Variance of likelihood value} \\ \cline{2-5}  
				         & \multicolumn{2}{c|}{Frame 5} & \multicolumn{2}{c|}{Frame 10} \\ \cline{2-5} 
				         & Fixed & Adaptive & Fixed & Adaptive \\ 
				         & 3D-DCT & 3D-DCT & 3D-DCT & 3D-DCT \\ \cline{1-5}
UCSD ped1 & 0.00112 & 0.00407 & 0.00136 & 0.00143 \\ \cline{2-5}
UCSD ped2 & 0.00193	& 0.00498 & 0.00193 & 0.0035 \\ \cline{2-5}
UMN & 0.00157 & 0.00484 & 0.00208 & 0.00396 \\ \cline{2-5}
CHUK-Avenue & 0.00128 & 0.00479	& 0.00227 & 0.00538 \\ \cline{2-5}
ShanghaiTech & 0.00098 & 0.00405 & 0.00159 & 0.00319 \\ \cline{1-5}
\end{tabular}}
\label{tab:tab5}
\end{table}
\begin{table}
  \centering
  \caption{Table showing paired t-test statistics on AUC over MOF-OF and MOF-saliency paired experiments}
  \scalebox{0.75}{
    \begin{tabular}{rrrrr}
          &       &       &       &  \\
\cmidrule{2-4}    \multicolumn{1}{r|}{} & \multicolumn{1}{c|}{\multirow{2}[4]{*}{Dataset}} & \multicolumn{2}{c|}{p-value} &  \\
\cmidrule{3-4}    \multicolumn{1}{r|}{} & \multicolumn{1}{c|}{} & \multicolumn{1}{c|}{MOF-OF pair} & \multicolumn{1}{c|}{MOF-saliency pair} &  \\
\cmidrule{2-4}    \multicolumn{1}{r|}{} & \multicolumn{1}{c|}{USCD PED1} & \multicolumn{1}{c|}{$5.44 \times 10^{-4}$} & \multicolumn{1}{c|}{$3.82\times 10^{-11}$} &  \\
\cmidrule{2-4}    \multicolumn{1}{r|}{} & \multicolumn{1}{c|}{UCSD PED2} & \multicolumn{1}{c|}{0.0491} & \multicolumn{1}{c|}{0.0012} &  \\
\cmidrule{2-4}    \multicolumn{1}{r|}{} & \multicolumn{1}{c|}{UMN } & \multicolumn{1}{c|}{0.0284} & \multicolumn{1}{c|}{0.0035} &  \\
\cmidrule{2-4}          &       &       &       &  
    \end{tabular}}%
  \label{tab:tab3}%
\end{table}%
\subsection{Results on UMN dataset}
On the UMN datset, the methods selected for comparison include the pure OF~\cite{Mehran_2009}, SF~\cite{Mehran_2009}, SRC~\cite{Cong_2013}, Interaction energy potentials (IEP)~\cite{Cui_2011}, chaotic invariants (CI)~\cite{Wu_2010}, local aggregates (LA)~\cite{Saligrama_2012}, Hierarchical-MDT variations with CRF filters (H-MDT-CRF)~\cite{Weixin_Li_2014}, (OADC-SA)~\cite{Yuan_Yuan_2015} and \cite{Biswas_2016}. In figure~\ref{fig:figure10}(c), a comparison of several frame-level ROC curves are shown on the entire UMN dataset for different competing methods. In table~\ref{tab:tab2}, we report the respective AUC values of our method for 3 different scenes as 0.988, 0.978, 0.984. Our AUC value on the whole UMN dataset is 0.982. The table clearly indicates that our results, without any form of training, are very much comparable with other competing methods which have used some form of training. For example, our method performed best (including win over \cite{Biswas_2016}) on Scene 2. Once again, we show qualitative response of our model for some of the UMN video sequences in figure~\ref{fig:figure11}(c).

We have performed comparative study with other saliency maps in the UMN dataset also (see figure~\ref{fig:figure16}). In this dataset, our adopted saliency detection method \cite{Gangapure_2018} yielded a slightly higher AUC value of 0.9823 over that of RWRV~\cite{Kim_2013} with an AUC of 0.9702. The AUC value of 0.9896 from SP-Liu~\cite{Liu_2014} is found to be quite comparable.  
\begin{table}
\centering
\caption{Performance measure showing AUC on frame-level anomaly for different frameworks over CHUK-Avenue and ShanghaiTech Campus dataset}
\scalebox{0.75}{
\begin{tabular}{ rrrrr }
\cmidrule{2-4}    \multicolumn{1}{r|}{} & \multicolumn{1}{c|}{\multirow{2}[4]{*}{Method}} & \multicolumn{2}{c|}{Dataset} &  \\
\cmidrule{3-4}    \multicolumn{1}{r|}{} & \multicolumn{1}{c|}{} & \multicolumn{1}{c|}{CHUK-Avenue} & \multicolumn{1}{c|}{ShanghaiTech Campus} &  \\
\cmidrule{2-4}
	& Conv-AE~\cite{Hasan_2016} & 0.745 & 0.6085 \\ \cmidrule{2-4}
	& Del \textit{et al.}~\cite{Del_Giorno_2016} & 0.783 & NA \\ \cmidrule{2-4}
	& TSC~\cite{Luo_2017} & 0.80568 & 0.6794 \\ \cmidrule{2-4}
	& sRNN~\cite{Luo_2017} & \textbf{0.8171} & 0.68 \\ \cmidrule{2-4}
	& \textbf{Ours} & 0.8099 & \textbf{0.768} \\ \cmidrule{2-4}
\end{tabular}}
\label{tab:tab6}
\end{table}    
\subsection{Results on CHUK-Avenue dataset}
On CHUK-Avenue dataset, Conv-AE~\cite{Hasan_2016} and Del \textit{et al.}~\cite{Del_Giorno_2016} have earlier achieved state-of-the-art performances. Recently published methods TSC~\cite{Luo_2017} and sRNN~\cite{Luo_2017} have now performed better than the previous methods. We have compared our performance with all these methods as shown in table~\ref{tab:tab6}. On this dataset, our model has yielded an overall AUC value of 0.8099. It ranks second by outperforming Conv-AE~\cite{Hasan_2016}, Del \textit{et al.}~\cite{Del_Giorno_2016} and TSC~\cite{Luo_2017} and by slightly falling behind sRNN~\cite{Luo_2017}. Please note that although the method developed by Del \textit{et al.} did not used any kind of labeled training data but yet have trained a logistic regression model based on distribution.

\subsection{Results on ShanghaiTech Campus dataset}
ShanghaiTech Campus dataset is a newly introduced relatively more complex datset with much diverse lightning and viewing conditions. On this dataset, we have achieved best results with an AUC value of 0.768. Here, we have outperformed Conv-AE~\cite{Hasan_2016}, TSC~\cite{Luo_2017} and sRNN~\cite{Luo_2017} datasets. As shown in table~\ref{tab:tab6}, the second best performance reported on this dataset was by sRNN~\cite{Luo_2017} with an AUC value of 0.68. Our AUC value of 0.768 surpasses the second best method by 8.8\%. 
\subsection{Execution Times}
Computation time plays a significant role in video surveillance. As reported in~\cite{Yuan_Yuan_2015}, the computational cost per frame for MDT~\cite{Mahadevan_2010} is 25 sec on a standard platform of 3 GHz processor. For SRC~\cite{Cong_2013} the average time spent on each frame is 3.8 sec and 0.8 sec for UCSD and UMN dataset respectively. STMC~\cite{Yang_Cong_2013} takes  1.2 sec to process a UCSD dataset frame. H-MDT-CRF~\cite{Weixin_Li_2014} reported an average time of 0.67 sec on UCSD \textit{ped1} dataset. Please note that the execution times for the above methods do not take into account the time required for training. Our method, which does not involve any training phase, required on average 2.17 sec and 3.25 sec to process a frame on UCSD \textit{ped1} and \textit{ped2} dataset respectively. A slight increase in the average computation time on UCSD \textit{ped2} is due to slightly increased resolution. It is to be noted that such computation time have been achieved without any code optimization or GPU acceleration. All implementations were done on MATLAB 2016a version installed on PC with 64-bit windows OS  with the processor speed of 3 GHz and 8 GB RAM. 
\section{Conclusion}
\label{sec:conclude}
In this paper, we proposed a training-less paradigm for anomaly detection which is self-adjustable in nature. Our model is capable of determining anomaly on-the-fly without requiring any form of prior training. This is quite useful for surveillance in areas from where availability of data cannot be guaranteed. We have achieved comparable performance with several state-of-the art methods on publicly available UCSD, UMN, CHUK-Avenue, ShanghaiTech datasets. In future, an extension of our work would be to effectively demarcate video frames with different degrees of threat as a convenience for faster search useful for long duration surveillance videos.


\bibliographystyle{abbrv}
\bibliography{BibFILE}

\end{document}